\documentclass[letterpaper, 10 pt, conference]{ieeeconf}  

\IEEEoverridecommandlockouts                              

\usepackage{graphics} 
\usepackage{subfigure}
\usepackage{caption}
\usepackage{epstopdf}
\usepackage{epsfig} 
\usepackage{amsmath} 
\usepackage{amssymb}  

\usepackage{comment}

\let\labelindent\relax
\usepackage{enumitem}

\usepackage{stfloats}

\usepackage{multirow}
\usepackage{makecell}

\usepackage{booktabs}

\title{\LARGE \bf
	Dynamic Movement Primitive based Motion Retargeting \\for Dual-Arm Sign Language Motions}

\author{Yuwei Liang$^{1}$, Weijie Li$^{1}$, Yue Wang$^{1}$, Rong Xiong$^{1,\ast}$, Yichao Mao$^{2}$, Jiafan Zhang$^{2}$
	\thanks{$^{1}$The authors are with Robotics Laboratory, Institute of Cyber-Systems and Control, Zhejiang University, China. 
		{\tt\small \{lyw.liangyuwei, liweijie, ywang24, rxiong\}@zju.edu.cn}}%
	\thanks{$^{2}$ABB Corporate Research Center, China.}
	\thanks{$^{*}$Corresponding author}%
	\thanks{This work was supported in part by the National Nature Science Foundation of China (Grant No. U1609210) and in part by Science and Technology Project of Zhejiang Province (Grant No. 2019C01043).}
}

\begin{document}

	\maketitle
	\thispagestyle{empty}
	\pagestyle{empty}

	\begin{abstract}
		
		We aim to develop an efficient programming method for equipping service robots with the skill of performing sign language motions. This paper addresses the problem of transferring complex dual-arm sign language motions characterized by the coordination among arms and hands from human to robot, which is seldom considered in previous studies of motion retargeting techniques.
		In this paper, we propose a novel motion retargeting method that leverages graph optimization and Dynamic Movement Primitives (DMPs) for this problem. We employ DMPs in a leader-follower manner to parameterize the original trajectories while preserving motion rhythm and relative movements between human body parts, and adopt a three-step optimization procedure to find deformed trajectories for robot motion planning while ensuring feasibility for robot execution.
		Experimental results of several Chinese Sign Language (CSL) motions have been successfully performed on ABB's YuMi dual-arm collaborative robot (14-DOF) with two 6-DOF Inspire-Robotics' multi-fingered hands, a system with 26 DOFs in total.
		
	\end{abstract}


	\section{Introduction}
	
	Service robots are gaining an increasing interest in recent years due to the ability to interact and communicate with humans. Nowadays, service robots interact with people mainly through visual or vocal means, e.g. a graphical monitor or synthetic voices, and there's no efficient way of communication for the hearing-impaired people. For service robots to assist these people, the ability to communicate via sign language is required.
	
	Previous works have developed various sign language robots. Studies that consider only the design and development of robotic hands, such as Dexter hand\cite{dexterous_hand} or Project Aslan's 3D-printed robotic arm\cite{proj_aslan}, can only implement fingerspelling, i.e., representing alphabet letters using only hands, which is inefficient and not practical for daily use. 
	Signs more commonly used in daily conversation often incorporate complex movements of both arms, and thus require the exploitation of dual-arm robots or humanoids. A dual-arm robot capable of Chinese Sign Language (CSL) was proposed in \cite{zhengda_undergrad}, but the vocabularies were limited to signs represented solely by \textbf{\emph{static signs}} and all the motions were manually programmed into the robot.
	Other studies implement \textbf{\emph{dynamic signs}} on humanoids for different research purposes, such as RASA\cite{RASA_PSL}, TEO\cite{TEO}, Nao H25 as well as Robovie R3\cite{kose1}\cite{kose2}\cite{kose3}\cite{kose4}. However, since these works focus on the design of robotic systems, social investigation, or education, they did not discuss the programming method or merely adopted manual programming.
	Honda company and Toshiba Corporation have presented appealing demonstrations of sign language on their own commercial humanoids, ASIMO\cite{ASIMO} and Aiko Chihira\cite{aiko_chihira}. But unfortunately, there are no formal research papers or reports as to how they program the motions into the robots.
	
	\newcommand{\chei}{0.1}
	\begin{figure}[t]
		\vspace{3mm}
		\centering
		\includegraphics[width=\linewidth]{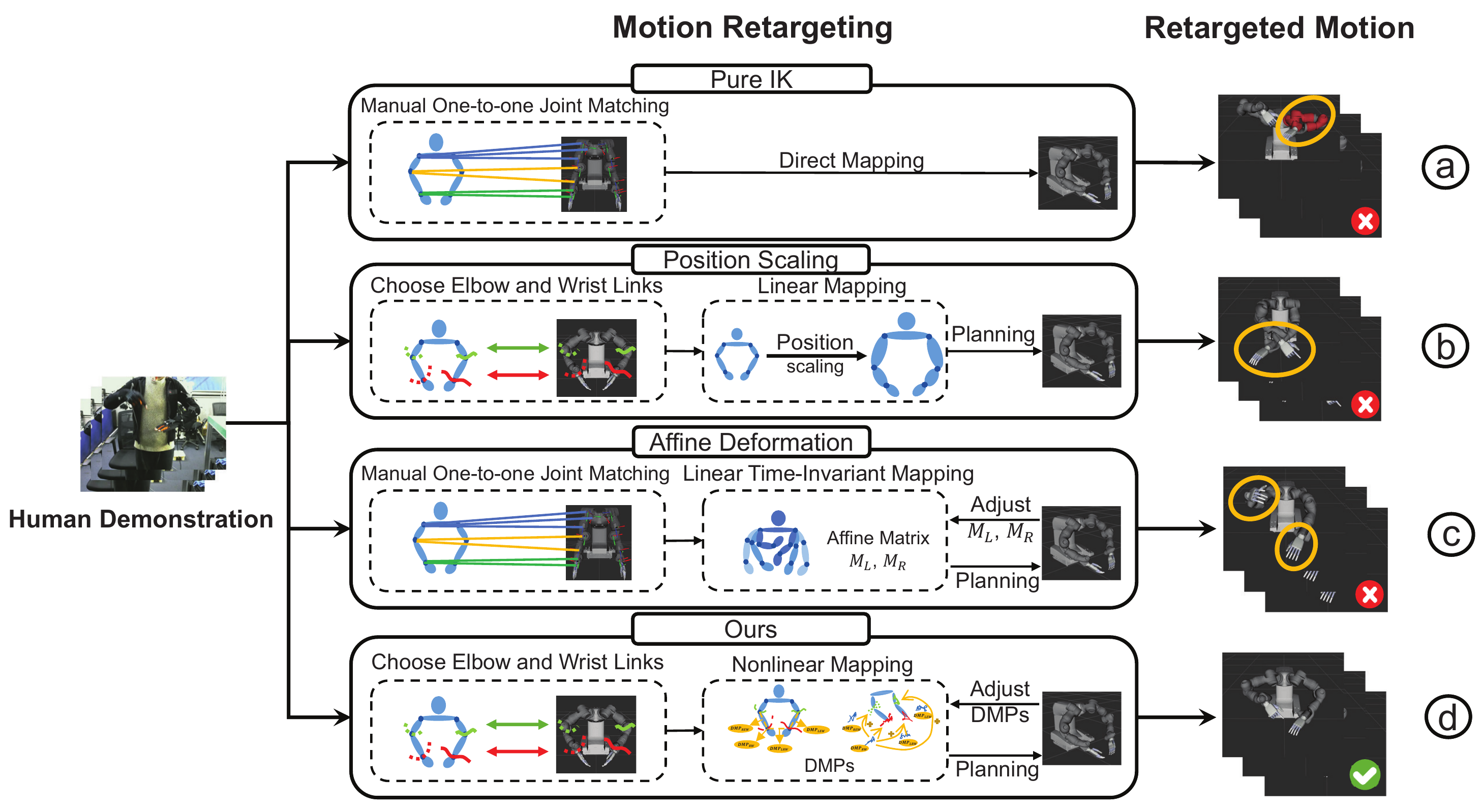}
		\caption{Retarget human demonstrated sign language motions onto a 26-DOF dual-arm robot using different methods. $(a)$ Infeasible motion. $(b)$ Misalignment between wrists. $(c)$ Inaccurate wrist orientation. $(d)$ Our method.}
		\vspace{-6mm}
	\end{figure}
	
	From these works, we see that \textbf{\emph{dynamic signs}} are necessary for effective communication, and manual programming or kinesthetic teaching is tedious and inefficient on account of the vast amount of words and various types of sign language.
	To efficiently program a robot with sign language motions, a technique called Motion Retargeting\cite{retargeting:gleicher} can be used to adapt human motions to robots. To cope with dual-arm sign language motions, we need a motion retargeting method that can effectively address the following problems:
	
	\begin{enumerate}[leftmargin=*]
		\item 
		\textbf{Differences in body structure and size.} This is the primary concern for motion retargeting and is more vital for sign language motions, which should be handled with proper modeling of dual-arm movements.
		
		\item
		\textbf{Dual-arm coordination.} Sign language motions convey semantic information through specific spatial and temporal relationship between movements of arms and hands, which should be preserved for the retargeted motions to be understood correctly. 
		
		\item
		\textbf{Motion similarity.} Directly using human demonstrations as reference does not consider differences in link lengths and workspace size. We need a more reasonable way to ensure similarity while allowing for slight deformation to the original trajectories.
		
		\item
		\textbf{Feasibility constraints.} Retargeted motions are supposed to be feasible for robot execution. Feasibility constraints and collision avoidance should be taken into account, especially for sign language motions involving coordination between body parts.
	\end{enumerate}
	
	In this paper, we propose a novel motion retargeting method that jointly addresses the above problems by combining Dynamic Movement Primitives (DMPs) \cite{DMP} and graph optimization\cite{g2o}. We first introduce an effective parameterization of arm state using wrist pose and elbow position to deal with differences in body structure. 
	Then, by learning DMPs in a leader-follower manner, motion rhythm as well as coordination pattern, i.e., relative movements between elbows and wrists, can be effectively retained for new trajectories generated by DMPs. 
	With the generalization ability of DMPs, we adopt optimization to search for deformed reference trajectories that can be well tracked by the robot regarding feasibility constraints and collision avoidance. The proposed method is validated on ABB's YuMi 14-DOF dual-arm collaborative robot with Inspire Robotics' 6-DOF multi-fingered hands, a system of 26 DOFs. We also conduct comparison experiments with other motion retargeting methods and the results show that our method can better handle dual-arm motions that involve coordination between both arms.

	\section{Related work}
	
	Motion retargeting has been widely investigated for decades in the field of computer graphics and robotics for transferring motions from one subject to another, where the source and target subjects are allowed to differ in body structure, mechanical properties, and size. In terms of robotics, physical constraints imposed by robot dynamics and collision avoidance are also considered to make sure the retargeted motions are feasible for robot execution. 
	
	Previous studies have proposed various motion retargeting methods for different applications, such as humanoid dancing\cite{retargeting:hrp_dancing}, robotic hand for dexterous manipulation\cite{retargeting:grasping}, and redundant manipulator making golf swing\cite{retargeting:golf_swing}. Researches on different types of robots impose different task requirements, but the fundamental routines are quite similar and can be roughly split into two parts - Motion Mapping and Constraints Handling. 
	The first part converts human demonstrations into a robot's reference joint trajectories, taking account of differences in body structure and size, during which motion similarity is the primary goal and the results are allowed to be infeasible. 
	In the second part, the reference joint trajectory will be processed regarding different task constraints as well as feasibility constraints, usually through optimization algorithms\cite{retargeting:gleicher}\cite{retargeting:tro_optim}. 
	
	According to the way of mapping human trajectories to robot's reference trajectories, motion retargeting methods can be categorized into two groups - Direct Mapping and Motion Modification. 
	
	\textbf{Direct Mapping.} Methods belonging to the group of direct mapping utilize the original human demonstrations without any modifications, either by tracking the originally recorded trajectories or using human Inverse Kinematics (IK) results. 
	Direct use of human joint trajectories is proposed in \cite{retargeting:sword_huangqiang}, but cannot cope with different structure or size since only joint angles are provided, and it requires pre-defined joint correspondence between robot and human.
	Marker position trajectories recorded by motion-capture systems can also be used to calculate the corresponding human joint trajectories via IK and incorporated into the similarity measure\cite{retargeting:hrp_dancing}\cite{retargeting:grasping}\cite{retargeting:online}, but still may not be suitable for a robot with different structure or size to follow. 
	In addition, a novel method\cite{retargeting:depth}\cite{retargeting:rgbd} converts human postures to robot joint angles by fitting a personalized 3D Human-Robot parametric model to captured point cloud from depth camera, which can deal with structure difference but only applies to simple postures due to occlusion problem.
	It can be seen that methods relying on direct mapping do not address differences in body structure and size, and only perform well on ``open-loop" motions that do not have strict requirements on the interaction between different body parts, e.g. dancing\cite{retargeting:hrp_dancing}, Kungfu ``Sword"\cite{retargeting:sword_huangqiang}, and waving arm\cite{retargeting:depth}.
	
	\textbf{Motion Modification.} To cope with differences in body structure and size, it's more reasonable to apply modifications to the original human demonstration so that it's more suitable for robot configuration and size. 
	Based on this idea, a natural thought is to apply scaling on the position data of, e.g. motion capture markers\cite{retargeting:taiji_asimo}, feature points extracted from depth image\cite{retargeting:taiji_asimo2}, or joints of interest\cite{retargeting:JOI}, according to link lengths or link orientation. This approach has the limitation that difference in link lengths results in position shifting on the end-effectors(hands) or intermediate joints, which could break the relative motions between different body parts.
	An optimization-based method\cite{retargeting:tro_optim} utilizes geometric parameter identification to avoid manually setting position scaling ratios by using a parametric model with adjustable link lengths and uses marker position error as the similarity measure. However, relative movements between different body parts are not explicitly addressed and there's no guarantee for preserving them during the morphing process.
	Apart from position scaling, a method\cite{retargeting:golf_swing} adopts homogeneous transformation for motion modification and employs Stochastic Optimization of the Embodiment Mapping (ISOEMP) to search for optimal shape and location of the human demonstration. This approach only applies to a single trajectory and cannot cope with multiple coordinated trajectories simultaneously. 
	Motion modification can also be done in joint space other than Cartesian space, such as applying affine transformation on human joint trajectories\cite{retargeting:hujin_dmp}\cite{retargeting:hujin_phdthesis}\cite{retargeting:affine}. Since modifications are done on joint trajectories, joint matching is required and differences in body structure and size are still not handled. 
	
	Recently, a deep learning-based approach\cite{retargeting:autoencoder} for character animation proposes the use of a parametric autoencoder to produce similar motions for skeletons with different link lengths. However, this method requires the source and target skeletons to share the same structure and essentially converts the problem of similarity measure into the problem of constructing datasets for training the parametric autoencoder.
	
	To the best of our knowledge, there is not yet any study that explicitly addresses the coordination between trajectories of multiple body parts during motion retargeting, which is usually encountered in more complex dual-arm motions, such as sign language motions, dual-arm assembly tasks.
	
	\section{Method}
	
	In this section, we present a novel motion retargeting method that leverages graph optimization and Dynamic Movement Primitives (DMPs) to transfer human demonstrated complex dual-arm motions to a robot.
	The proposed method consists of three parts: modeling of human arm movements, learning DMPs in a leader-follower manner from human demonstrations, and optimization of initial and goal positions of DMPs. We now elaborate on them in the following subsections.

	\subsection{Modeling of Arm Movements}
	
\begin{figure}[htb]
		\centering
		\includegraphics[scale=0.35]{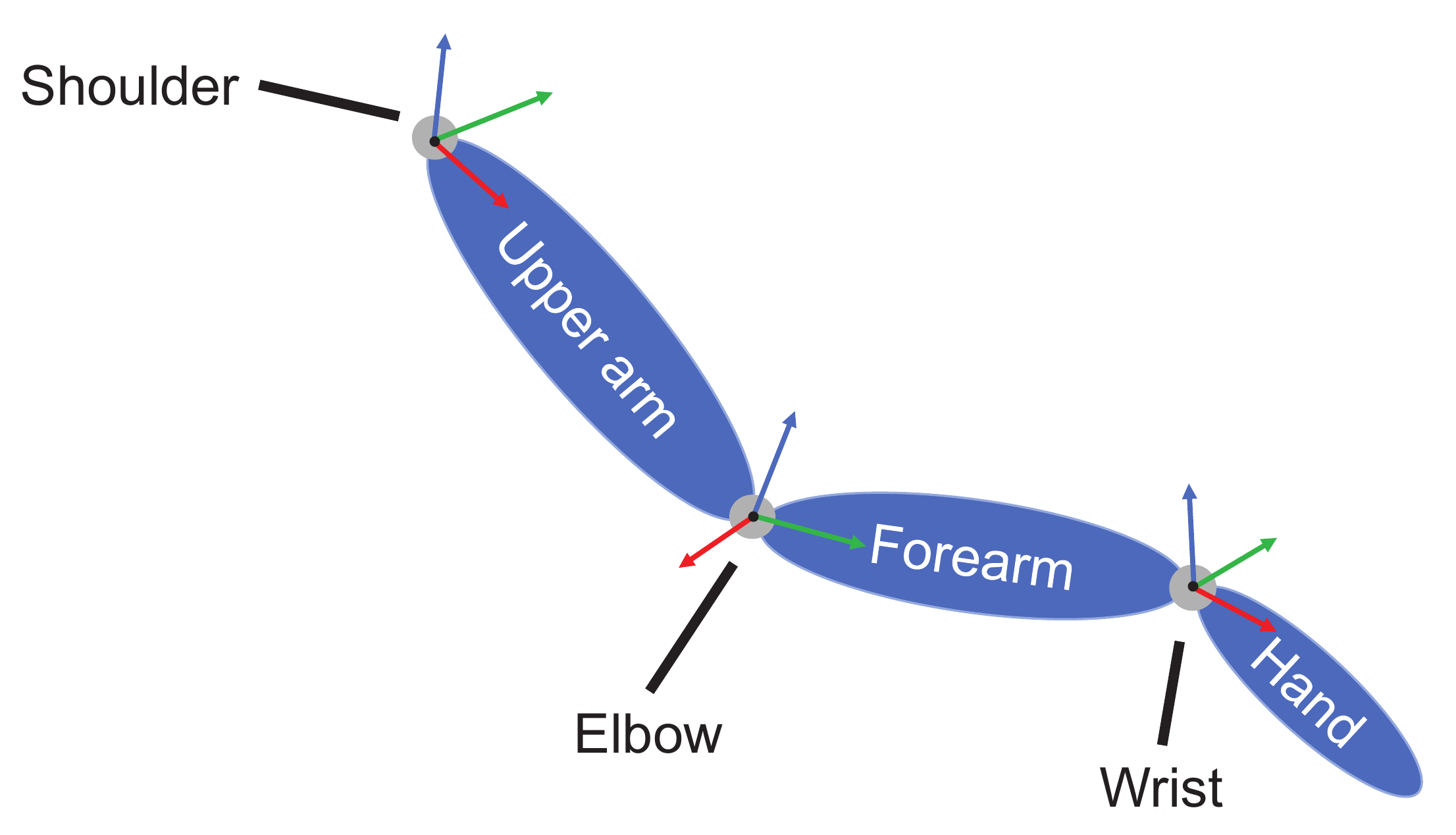}
		\caption{A simplified three-link arm model.}
		\label{arm_model}
		\vspace{-3mm}
	\end{figure}

	To avoid tedious, robot-specific joint matching and fully capture human-likeness, we choose to describe the arm state in Cartesian space with a simplified three-link model, treating the upper arm, forearm, and hand as separate rigid bodies, as is illustrated in Fig. \ref{arm_model}.
	\vspace{-3mm}

	Three free-floating rigid bodies in space have 18 DOFs. With the assumption that shoulder position, upper arm length, and forearm length are known, 9 constraints are introduced and reduce the system DOFs to 9. As previous researches\cite{retargeting:elbow_1}\cite{retargeting:elbow_2} indicate that elbow position affects anthropomorphism of a generated motion, we decide to model arm state with elbow position and wrist pose (position and orientation), using 9 independent variables in total.

	\subsection{Learning DMPs in a Leader-Follower Manner} 
	
	DMP\cite{DMP} is a dynamical system based imitation learning method, which consists of two parts: a transformation system (Eq. \ref{eq:dmp_transformation}) and a canonical system (Eq. \ref{eq:dmp_canonical}).
	\begin{equation}
		\tau\ddot{y}=K(g-y)-D\dot{y}+f, f=\frac{\sum_i{\phi_i\omega_i}}{\sum_i{\phi_i}}x(g-y_0)
		\label{eq:dmp_transformation}
	\end{equation}
	\begin{equation}
		\tau{\dot{x}}=-\alpha x, x(0)=1
		\label{eq:dmp_canonical}
	\end{equation}
	The transformation system is essentially a globally stable second-order dynamical system with a unique point attractor, added by a learnable nonlinear forcing term $f$, and guarantees convergence to the point $(0,g)$ from arbitrary initial state $(1,y_0)$ while preserving the path shape of input trajectory.
	The canonical system serves as an internal clock by providing a phase variable $x$ and helps in capturing the rhythm of the input trajectory.
	\begin{figure}[htb]
		\vspace{-2mm}
		\centering
		\includegraphics[scale=0.25]{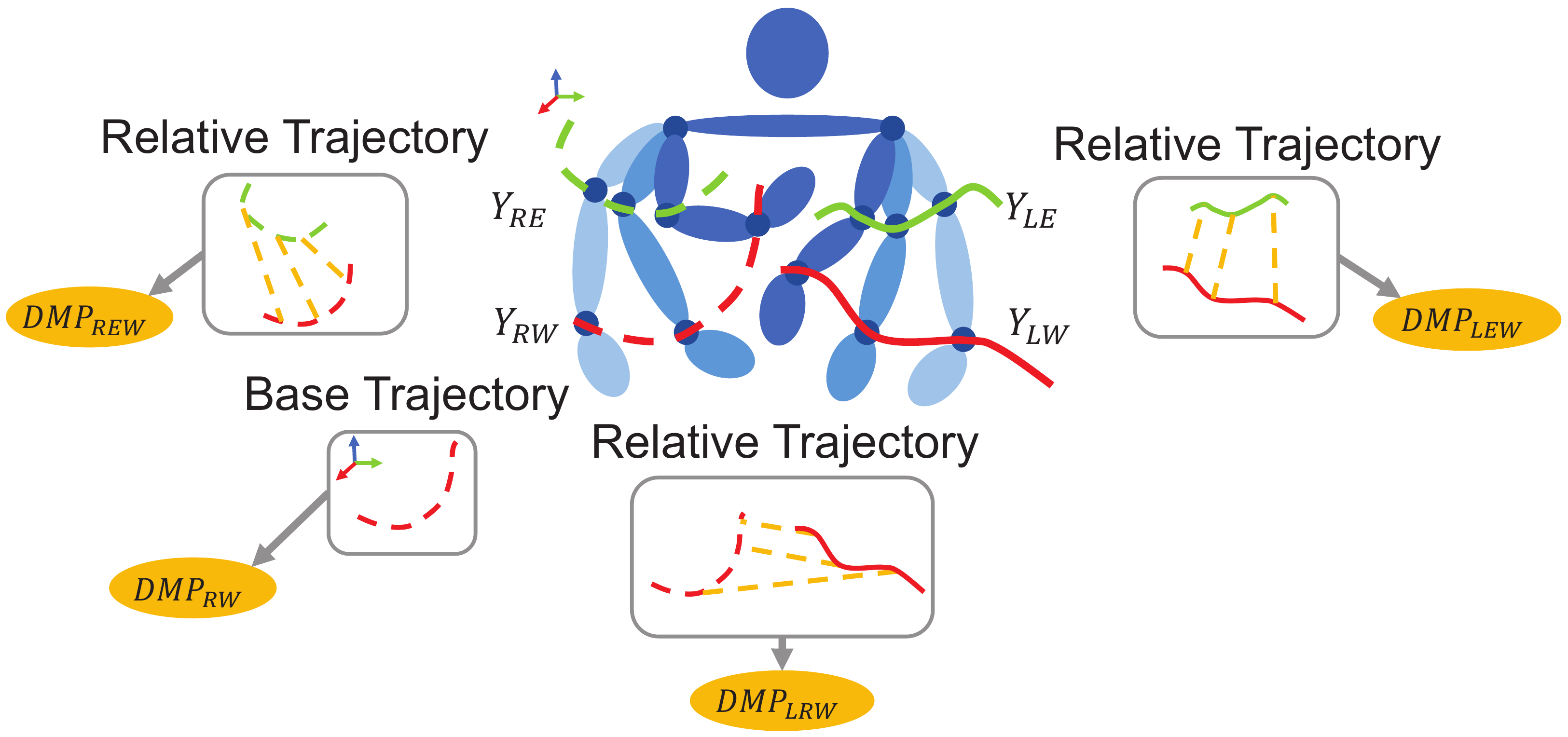}
		\caption{Learning DMPs in a leader-follower manner.}
		\label{dmp_learning}
		\vspace{-3mm}
	\end{figure}
	
	To preserve relative movements between elbows and wrists, we apply four DMPs in a leader-follower manner to learn from the motion-captured position trajectories, that is, one for right wrist trajectory which serves as the base trajectory ($Y_{RW}$), one for relative trajectory between left and right wrists ($Y_{LRW}$), and the rest for relative trajectories between elbows and wrists of the same arm ($Y_{LEW}$, $Y_{REW}$), as is illustrated in Fig. \ref{dmp_learning} and formulated in Eq. \ref{eq:four_dmps}. 
	\begin{equation}
		\begin{split}
			{DMP}_{RW} &\leftarrow Y_{RW}\\
			{DMP}_{LRW} &\leftarrow Y_{LRW} = Y_{LW} - Y_{RW}\\
			{DMP}_{LEW} &\leftarrow Y_{LEW} = Y_{LE} - Y_{LW}\\
			{DMP}_{REW} &\leftarrow Y_{REW} = Y_{RE} - Y_{RW}
		\end{split}
		\label{eq:four_dmps}
	\end{equation}
	Here $Y_{LW}$, $Y_{LE}$, $Y_{RW}$, and $Y_{RE}$ are position trajectories of wrists and elbows of both arms respectively. Note that orientation trajectories are not learned by DMPs but directly used as the reference since absolute orientation of two hands is not supposed to deviate much from the demonstration for a particular sign language motion. 
	
	Before learning DMPs, the motion-captured wrist position trajectories should be pre-processed according to human's and robot's hand lengths to cope with different hand size, which is not considered in our arm model.
	We apply translation to the wrist position along the direction to which the hands point according to human's and robot's hand lengths for the robot hand to reach the same virtual hand tip position, as is formulated in Eq. \ref{eq:wrist_trans}.
	\begin{equation}
		\begin{split}
			p_{VHT}&=p_{H,W} + R_{H,W}\cdot[0, 0, L_H]^T\\
			p_{R,W}&=p_{VHT} + R_{R,W}\cdot[0, 0, -L_R]^T
		\end{split}
		\label{eq:wrist_trans}
	\end{equation}
	$p_{VHT}$ is the virtual hand tip position. $p_{H,W}$, $p_{R,W}$, $R_{H,W}$, $R_{R,W}$, $L_H$ and $L_R$ are wrist positions, orientation and hand lengths of human and robot respectively.
	
\begin{figure}[htbp]
		\vspace{2mm}
		\centering
		\includegraphics[scale=0.2]{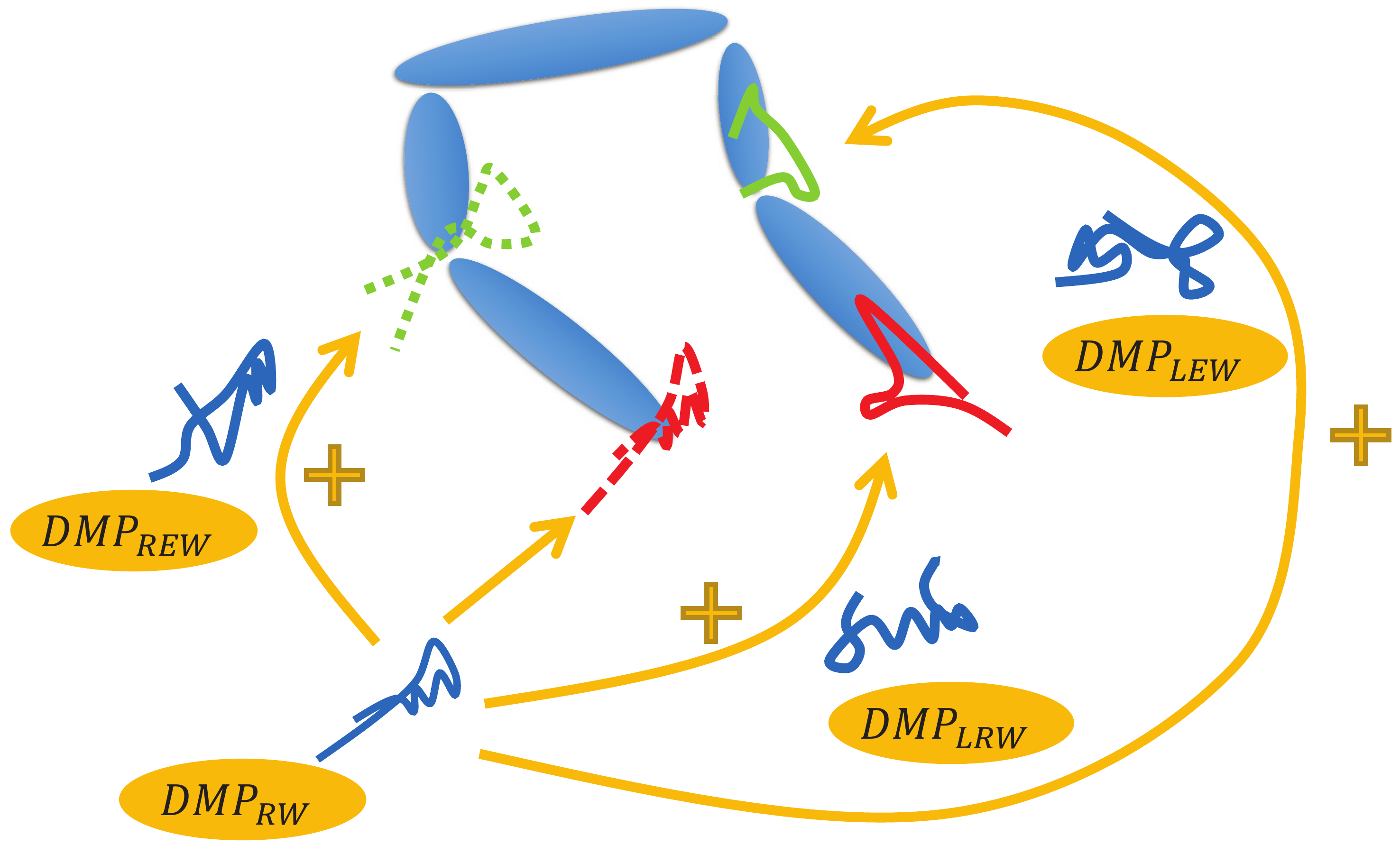}
		\caption{Superimpose the reproduced base and relative trajectories to make up reference trajectories for wrists and elbows of both arms.}
		\label{dmp_repro}
		\vspace{-6mm}
	\end{figure}

	After adjusting weights for DMPs using learning algorithms such as Locally Weighted Regression (LWR), four DMPs can be used later to reproduce new base and relative trajectories given new initial and goal positions, which are then superimposed to make up new trajectories for wrists and elbows of both arms, as is shown in Fig. \ref{dmp_repro}.

	\subsection{DMP-based Motion Retargeting Framework}
	
	DMP can be viewed as a parameterized trajectory $\hat{Y}_i(n) = DMP_i(S_i, G_i, n)$, $n=0, \dots, N$, with adjustable initial position $S_i$ and goal position $G_i$, which preserves the original path shape to some extent when generalized to new locations.
	This generalization property allows us to adopt a three-step optimization procedure to find deformed trajectories that satisfy robot constraints by treating initial and goal positions of DMPs as optimization variables.
	\begin{figure}[htbp]
		\centering
		\includegraphics[width=\linewidth]{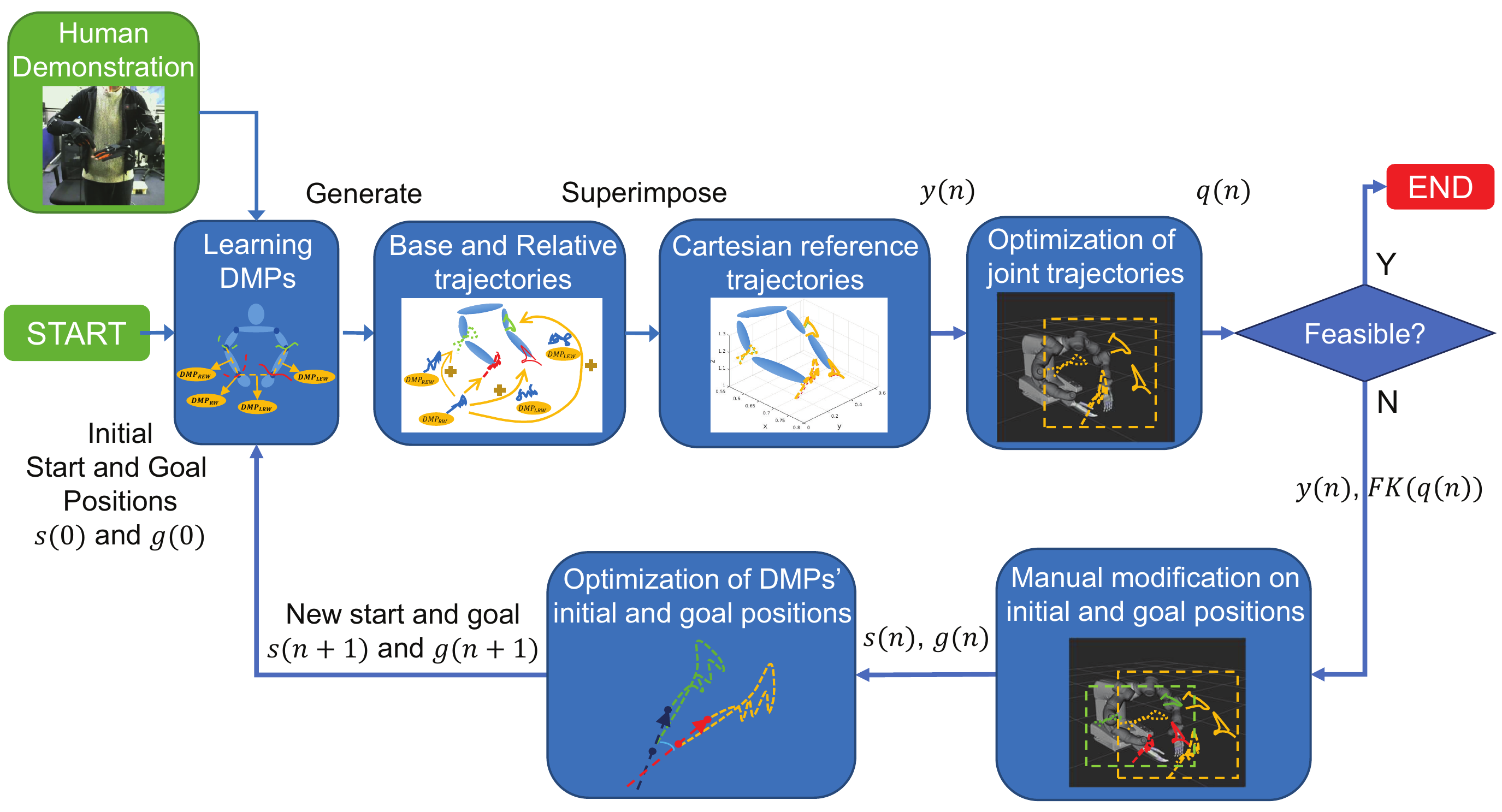}
		\caption{Three-step optimization framework for optimizing joint trajectories as well as initial and goal positions of DMPs.}
		\label{dmp_optim}
		\vspace{-4mm}
	\end{figure}

	As is illustrated in Fig. \ref{dmp_optim}, we first provide an initial set of $(S_i, G_i)$ for DMPs to generate one base trajectory and three relative trajectories, which can be superimposed to produce reference trajectories for elbows and wrists:
	\begin{equation*}
		\begin{aligned}
			& \hat{Y}_{RW} = DMP_{RW}(S_{RW},G_{RW}) \\
			& \hat{Y}_{LW} = DMP_{LRW}(S_{LRW},G_{LRW})+\hat{Y}_{RW} \\
			& \hat{Y}_{LE} = DMP_{LEW}(S_{LEW},G_{LEW})+\hat{Y}_{LW} \\
			& \hat{Y}_{RE} = DMP_{REW}(S_{REW},G_{REW})+\hat{Y}_{RW} 
		\end{aligned}
	\end{equation*}
	Given the produced position trajectories and the original orientation trajectories, the corresponding robot joint trajectory can be obtained by optimizing a tracking objective function and taking feasibility constraints into account, as formulated in the following least squares minimization problem,
	\begin{equation*} 
		\begin{aligned}
			\min_{Q=\{q_n\}_{n=0}^N} & e_{TRK}^T(Q,\hat{Z})e_{TRK}(Q,\hat{Z}) + e_{COL}^T(Q)e_{COL}(Q) \\
			& + e_{SMO}^T(Q)e_{SMO}(Q) + e_{PLT}^T(Q)e_{PLT}(Q) \\
		\end{aligned}
		\label{optim:q}
	\end{equation*}
	where $\hat{Z}=\{\hat{Y}_j,\hat{R}_j\}$, $j\in\{LW,RW,LE,RE\}$ contains both position and orientation goals, $e_{TRK}(Q,\hat{Z})$ is the tracking cost of position and orientation, $e_{COL}(Q)$ is the collision cost, $e_{SMO}(Q)$ is the smoothness cost, and $e_{PLT}(Q)$ is the position limit cost.
	
	After the robot joint trajectory is obtained, we move the reference position trajectories $\hat{Y}_i(n)$ closer to the actually tracked ones $Y_i(n)$ for ease of robot motion planning in the next round, by subtracting the average error between them from the initial and goal positions of the corresponding DMPs.
	\begin{gather*}
		\Delta Y_i = \frac{1}{N+1}\sum_{n=0}^{N}(Y_i(n)-\hat{Y}_i(n)) \\
		S_i \leftarrow S_i + \Delta Y_i, \quad G_i \leftarrow G_i + \Delta Y_i \\
		i \in \{RW, LRW, LEW, REW\}
	\end{gather*}
	
	Since manually modifying $\{S_i, G_i\}$ of DMPs could alter the relative movements of the original demonstration in an unwanted way, we devise three constraints for them to penalize unwanted deviation from the original values and optimize them regarding the constraints. One is the scaling cost (Eq. \ref{eq:scaling_cost}) which penalizes the change in scale of the whole trajectory by limiting the change of distance between $S_i$ and $G_i$. One is the orientation cost (Eq. \ref{eq:orien_cost}) that penalizes the change in orientation of the whole trajectory by limiting the direction of the vector pointing from $S_i$ to $G_i$. The last one is the cost of relative change (Eq. \ref{eq:rel_cost}) which is specifically for relative movements and penalizes the deviation from the original values directly. The cost functions are formulated as follows:
	\begin{equation}
		e_{SCL}(S_i, G_i) = max(\frac{ \lVert S_i-G_i \rVert}{\lVert \widetilde{S}_i - \widetilde{G}_i \rVert}, \frac{\lVert \widetilde{S}_i - \widetilde{G}_i \rVert}{ \lVert S_i-G_i \rVert})
		\label{eq:scaling_cost}
	\end{equation}
	\begin{equation}
		e_{ORI}(S_i, G_i) = acos(\frac{(G_i-S_i)^T(\widetilde{G}_i-\widetilde{S}_i)}{\lVert G_i - S_i \rVert \lVert \widetilde{G}_i - \widetilde{S}_i \rVert})
		\label{eq:orien_cost}
	\end{equation}
	\begin{equation}
		e_{REL}(S_i, G_i) = \lVert S_i - \widetilde{S}_i \rVert \lVert G_i - \widetilde{G}_i \rVert
		\label{eq:rel_cost}
	\end{equation}
	\noindent where $\widetilde{S}_i$, $\widetilde{G}_i$ are the initial and goal positions of base and relative trajectories, calculated from the original demonstration.
	We optimize the initial and goal positions of four DMPs by the following optimization problem:
	\begin{equation*}
		\begin{aligned}
			\min_{S,G} & \quad \sum_{i\in I} e_{SCL}^T(S_i,G_i)e_{SCL}(S_i,G_i) \\
			& + \sum_{i\in I} e_{ORI}^T(S_i,G_i)e_{ORI}(S_i,G_i) \\
			& + \sum_{j\in I_{rel}} e_{REL}^T(S_j,G_j)e_{REL}(S_j,G_j) 
		\end{aligned} 
		\label{optim:dmp}
	\end{equation*}
	\noindent where $S=\{S_i\}_{i\in I}$, $G=\{G_i\}_{i\in I}$ are initial and goal positions respectively. $I_{rel}=\{LRW,LEW,REW\}$ and $I = \{RW\} \cup I_{rel}$ are indices denoting base or relative trajectories.
	A new set of $(S_i, G_i)$ can be obtained after optimization, and would be used in the next round of optimization to generate new reference trajectories. The optimization procedure continues until the optimized joint trajectories are collision free and satisfy all feasibility constraints.

	\subsection{Implementation}
	
	We employ \emph{General Graph Optimization} (g2o) library\cite{g2o} for the optimization of joint trajectories and the initial and goal positions of four DMPs. 
	Since tracking of both wrist's and elbow's trajectories may not be satisfactory at the same time, we choose to prioritize the tracking goals of wrist and elbow by embedding nullspace formulation in the update rule of optimization.
	For collision avoidance, we adopt artificial potential field method with the use of robot jacobians, and also interpolate between adjacent path points to perform dense collision checking for approximating the effect of continuous collision checking.
	For hand motions, we simply use linear mapping on the finger joint angles measured by datagloves according to the range of motion for each joint, as is formulated in Eq. \ref{eq:linear_mapping}.
	\begin{equation}
		q_{r,i} = \frac{q_{h,i}-q_{h,imin}}{q_{h,imax}-q_{h,imin}}(q_{r,imax}-q_{r,imin}) + q_{r,imin}
		\label{eq:linear_mapping}
	\end{equation}

	\section{Experimental results}
	
	This section presents experimental results for our proposed motion retargeting method. First, to validate the effectiveness of our method, we conduct experiments on several Chinese Sign Language motions and quantitatively compare our results with other retargeting methods in terms of the percentage of feasible solutions and motion similarity. Then, we demonstrate the retargeted motions of different methods for perceptual evaluation. 
	
	\begin{table*}[htbp]
		\vspace{1mm}
		\centering
		\setlength{\tabcolsep}{2pt} 
		\renewcommand{\arraystretch}{1.2} 
		\begin{center}
			\caption{Frechet distances of absolute and relative trajectories.}
			\label{table:frechet_dists3}
			\begin{tabular}{l ccccccccc ccccccccc}	
				\hline
				& \multicolumn{9}{c}{Affine Deformation} & \multicolumn{9}{c}{Ours} \\
				\cmidrule(lr){2-10} \cmidrule(lr){11-19} 
				& LW & RW & LE & RE & LRW & LEW & REW & LWO & RWO & LW & RW & LE & RE & LRW & LEW & REW & LWO & RWO \\
				\hline
				$``baozhu"$ (firecracker) 		& \textbf{0.141} & 0.341 & \textbf{0.236} & 0.187 & 0.153 & 0.146 & 0.218 & 6.213 & 6.003 & 0.235 & \textbf{0.193} & 0.545 & \textbf{0.178} & \textbf{0.119} & \textbf{0.132} & \textbf{0.137} & \textbf{0.045} & \textbf{0.026} \\
				$``fengren"$ (sewing)		& 0.792 & 0.521 & 0.674 & 0.914 & 0.189 & 0.183 & 0.232 & 0.836 & 1.791 & \textbf{0.403} & \textbf{0.195} & \textbf{0.230} & \textbf{0.447} & \textbf{0.056} & \textbf{0.104} & \textbf{0.098} & \textbf{0.314} & \textbf{0.092} \\
				$``minzheng"$ (civil affairs) 	& 0.387 & 0.541 & 0.349 & 0.505 & 0.296 & \textbf{0.153} & 0.243 & 6.157 & 6.066 & \textbf{0.386} & \textbf{0.230} & \textbf{0.279} & \textbf{0.334} & \textbf{0.133} & 0.170 & \textbf{0.157} & \textbf{0.105} & \textbf{0.022} \\
				$``jieshou"$ (accept) 		& 0.464 & 0.535 & \textbf{0.524} & 0.678 & 0.198 & 0.124 & 0.163 & 6.120 & 6.020 & \textbf{0.373} & \textbf{0.207} & 0.649 & \textbf{0.539} & \textbf{0.050} & \textbf{0.082} & \textbf{0.151} & \textbf{0.018} & \textbf{0.012} \\
				$``shuan"$ (fasten)		& \textbf{0.376} & 0.539 & \textbf{0.685} & 0.585 & 0.246 & 0.178 & 0.194 & 4.567 & 6.018 & 0.519 & \textbf{0.172} & 0.842 & \textbf{0.161} & \textbf{0.130} & \textbf{0.121} & \textbf{0.157} & \textbf{0.103} & \textbf{0.028} \\		
				$``zhenli"$ (truth) 		& \textbf{0.260} & 0.543 & \textbf{0.642} & \textbf{0.328} & 0.231 & \textbf{0.139} & 0.217 & 4.887 & 6.014 & 0.398 & \textbf{0.149} & 0.990 & 0.431 & \textbf{0.131} & 0.151 & \textbf{0.161} & \textbf{0.026} & \textbf{0.016} \\
				\hline
			\end{tabular}
		\end{center}
		\vspace{-8mm}
	\end{table*}

	\subsection{Experimental setup}
	
	The proposed method is tested on a subset of words chosen from Chinese Sign Language that involve complex dual-arm interaction and coordination. The motions are recorded with a motion-capture system and datagloves, and then retargeted to ABB's YuMi dual-arm collaborative robot (14-DOF) with Inspire-Robotics' multi-fingered hands (6-DOF). Since YuMi robot has a different kinematic structure than humans or humanoid robots, our method exhibits independence of body structure in the experiments.
	
	We compare our method with three other retargeting methods. 
	One is a \textbf{PureIK method} in the group of Direct Mapping as described in Section \uppercase\expandafter{\romannumeral2}, which adopts a similar methodology as the work\cite{retargeting:hrp_dancing}. For our case, we employ nullspace control for tracking wrist pose and elbow position of the human demonstration. 
	Another one is an \textbf{affine transformation based method}\cite{retargeting:hujin_dmp}\cite{retargeting:hujin_phdthesis}\cite{retargeting:affine} belonging to the group of Motion Modification, which applies affine transformation on human joint trajectories, and uses human wrist and elbow trajectories as optimization objective to obtain the optimal affine transformation matrix. When manually determining joint correspondence between human and the robot, we find that the rotations of YuMi's wrist joint include pitch and roll, instead of pitch and yaw like humans, so we set the roll joint angle to 0 and leave it for optimization.
	The last one is based on \textbf{position scaling}\cite{retargeting:taiji_asimo}\cite{retargeting:taiji_asimo2}\cite{retargeting:JOI} with respect to shoulder positions according to link lengths.

	\subsection{Comparison results}
	
	We chose 15 Chinese Sign Language words and employed different retargeting methods to obtain retargeted motions.

	\textbf{Success Rates Evaluation.} In Table \ref{table:success_ratio}, we present the success rates by the percentage of retargeted motions that are feasible for robot execution.
	In the experiment, although the PureIK method tracks human wrist trajectories very well, it usually produces infeasible results, i.e. self-colliding motions, mostly because the original human motions are not suitable for a robot to replicate due to differences in body structure and size. The position scaling based method has the same problem and even induces misalignment between wrists' movements, which is shown in Fig. \ref{fig:fengren_1_snapshots}.
	\begin{table}[htbp]
		\vspace{-2mm}
		\setlength{\tabcolsep}{5pt} 
		\renewcommand{\arraystretch}{1.3} 
		\begin{center}
			\caption{Success rate of different retargeting methods.}
			\begin{tabular}{l cccc}
				\hline
				& \makecell[c]{PureIK\\\cite{retargeting:hrp_dancing}} & \makecell[c]{Position Scaling\\\cite{retargeting:taiji_asimo}\cite{retargeting:taiji_asimo2}\cite{retargeting:JOI}} & \makecell[c]{Affine\\\cite{retargeting:hujin_dmp}\cite{retargeting:hujin_phdthesis}\cite{retargeting:affine}} & Ours \\
				\hline
				Success rate & $0.0\%$ & $0.0\%$ & $40.0\%$ & $100.0\%$ \\
				\hline
			\end{tabular}
			\label{table:success_ratio}
		\end{center}
		\vspace{-6mm}
	\end{table}

	\textbf{Comparative Analysis in Motion Similarity.} To validate our method's ability to preserve path shape and relative movements, we compute Frechet distances of absolute (normalized to $[0,1]$) and relative trajectories, which is a commonly used quantitative measure for trajectory similarity. The smaller the Frechet distance between two trajectories is, the more similar their shapes are.
	Orientation trajectories of both wrists, denoted by $LWO$ and $RWO$ respectively, are also included for more complete comparison.
	We display some of the results in Table \ref{table:frechet_dists3}, where smaller Frechet distance values between resultant trajectories of two methods are highlighted in bold. We omit the PureIK and the Position Scaling methods since they did not yield any feasible results on our selected motions. 
	The Frechet distances of absolute position trajectories do not show much advantages of our method in preserving the path shape of the absolute trajectories other than that of right wrist. But the results of relative trajectories indicate that our method can better capture the original relative movements in the retargeted motions. More importantly, the retargeted orientation trajectories bear much closer resemblance to the original demonstration, which are important for generating understandable motions.

\begin{figure*}[htbp]
		\vspace{1.25mm}
		\centering
		\subfigure{
			\includegraphics[width=0.32\linewidth]{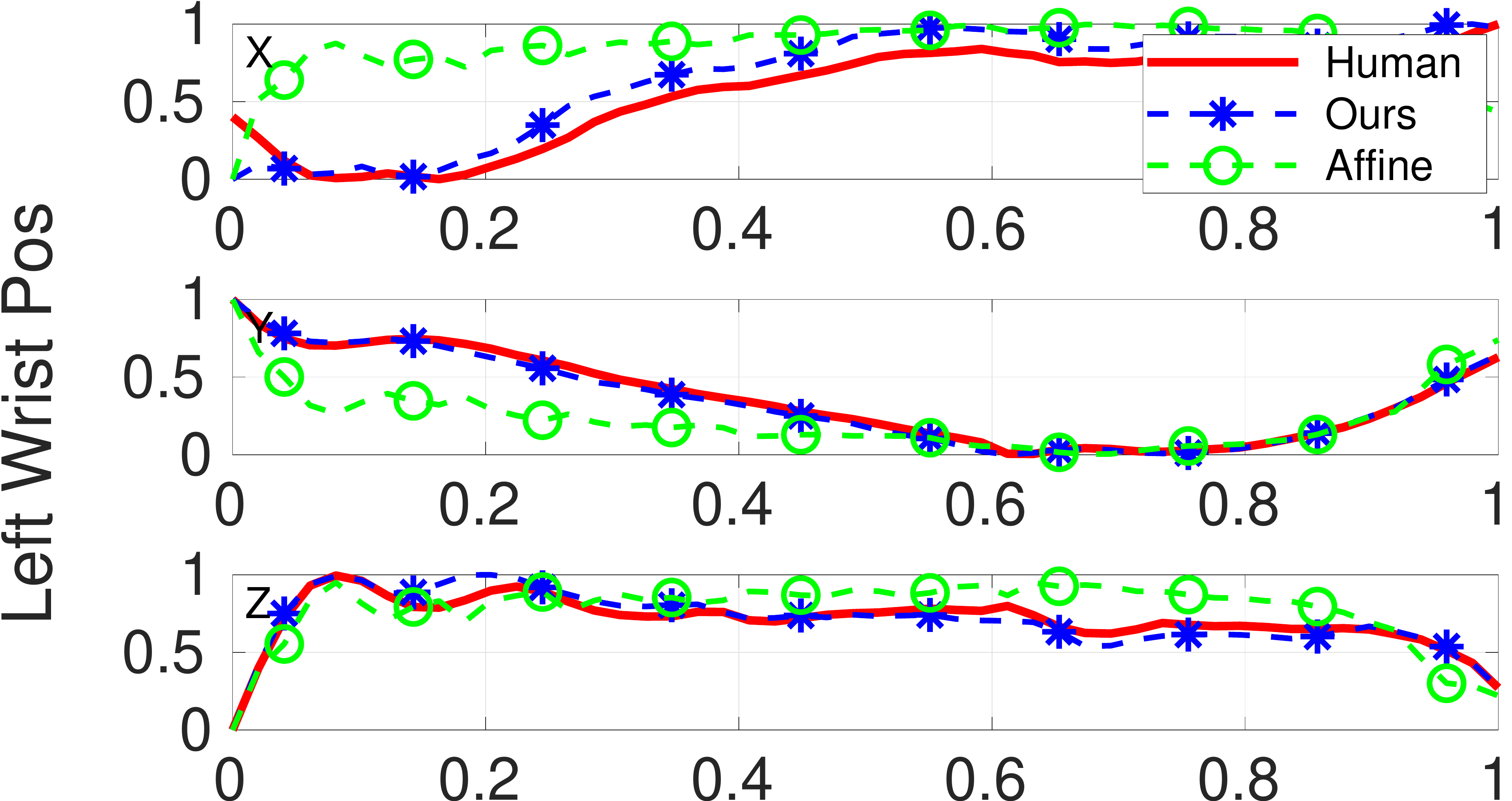}
			\includegraphics[width=0.32\linewidth]{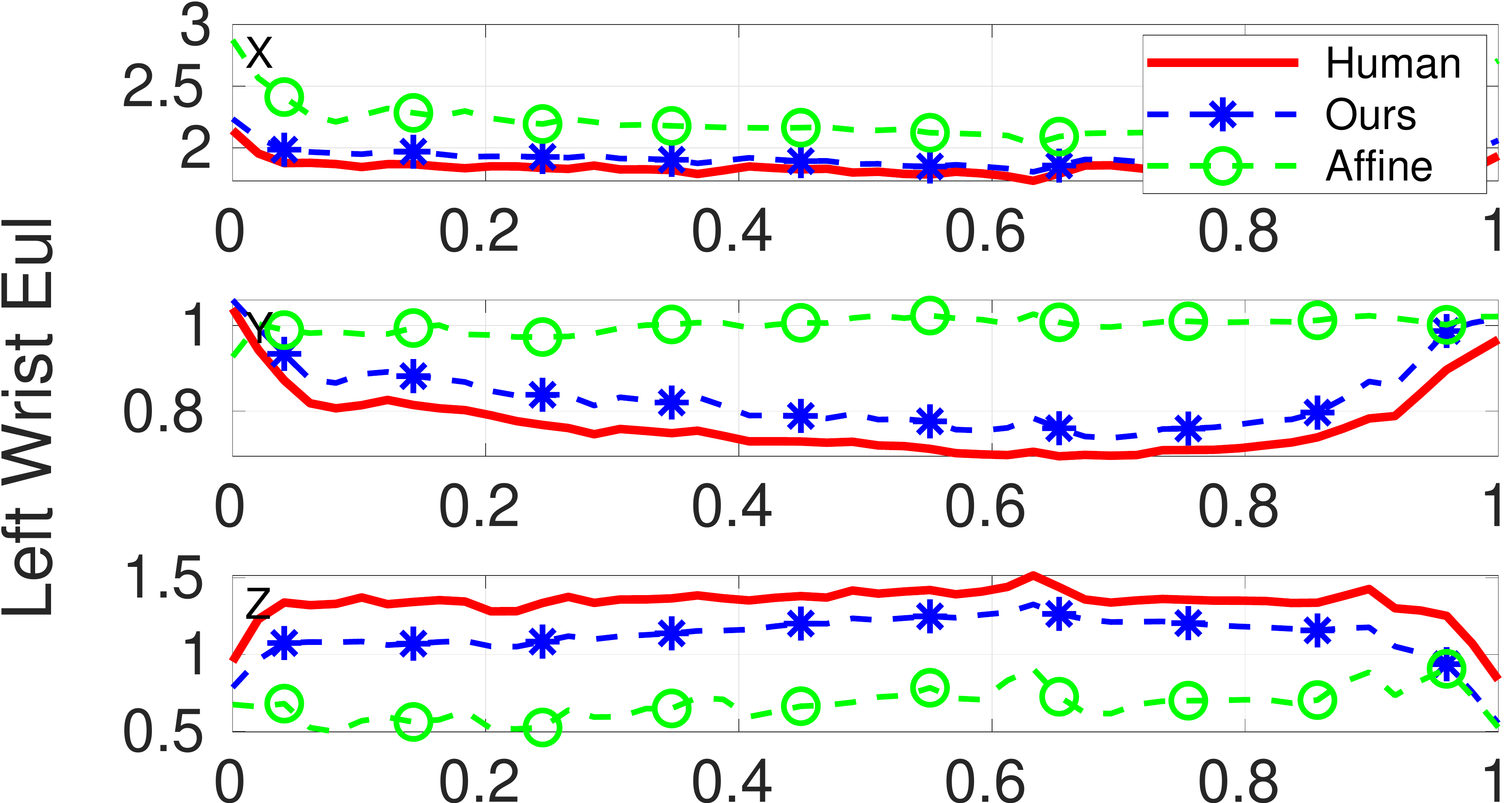}
			\includegraphics[width=0.32\linewidth]{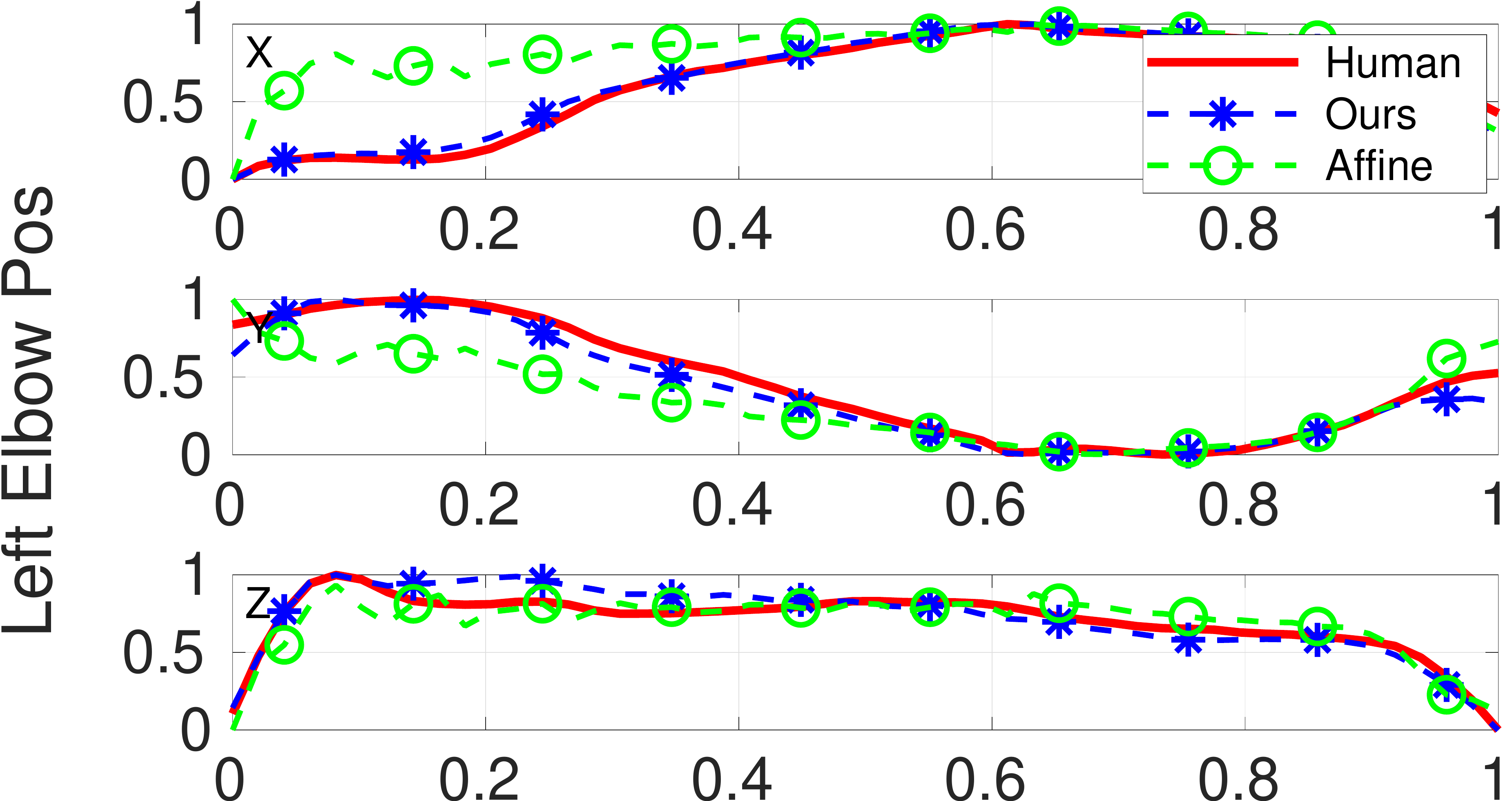}
			\label{fig:l_arm_fengren_1}
		} 
		\subfigure{
			\includegraphics[width=0.32\linewidth]{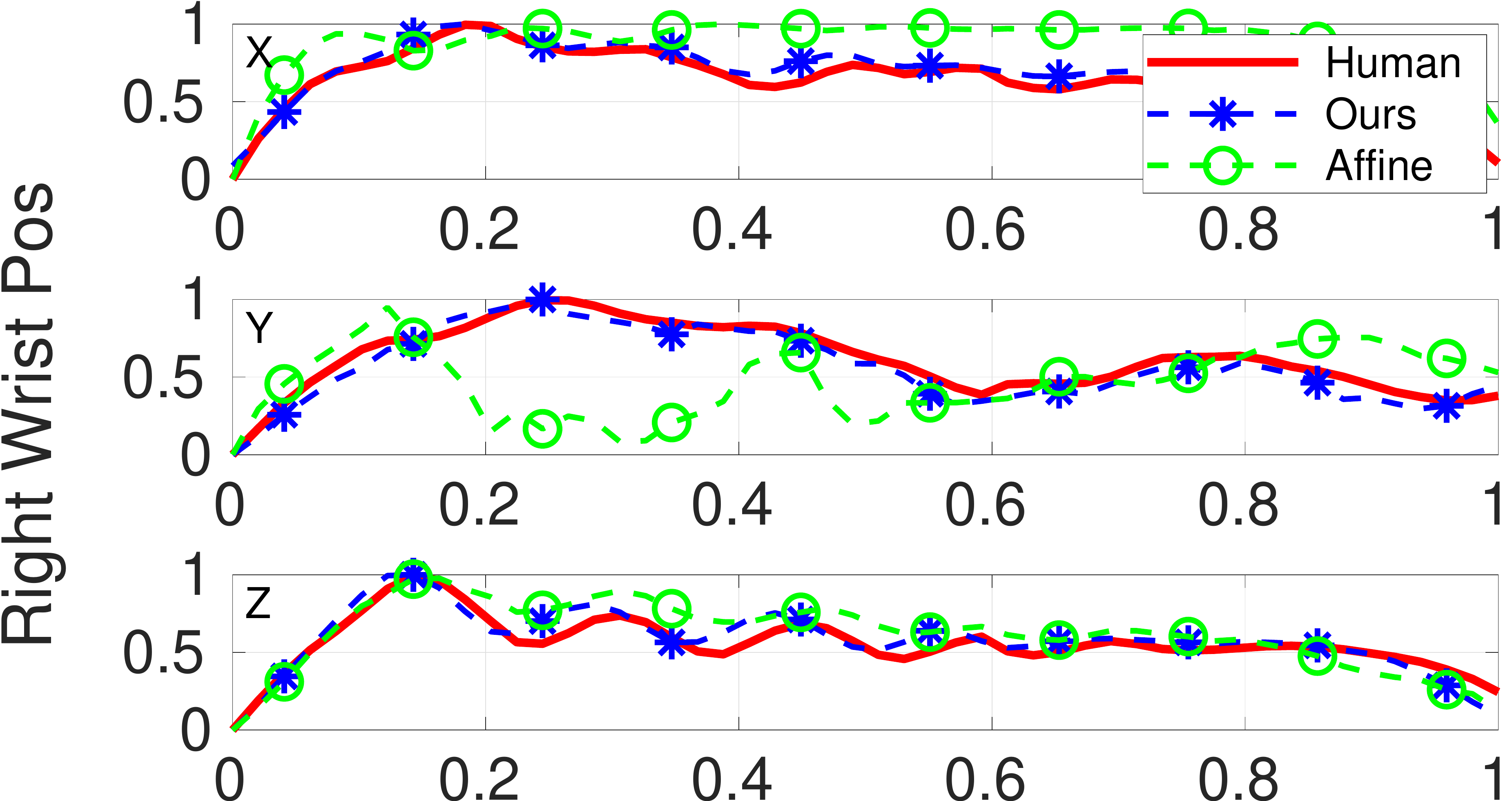}
			\includegraphics[width=0.32\linewidth]{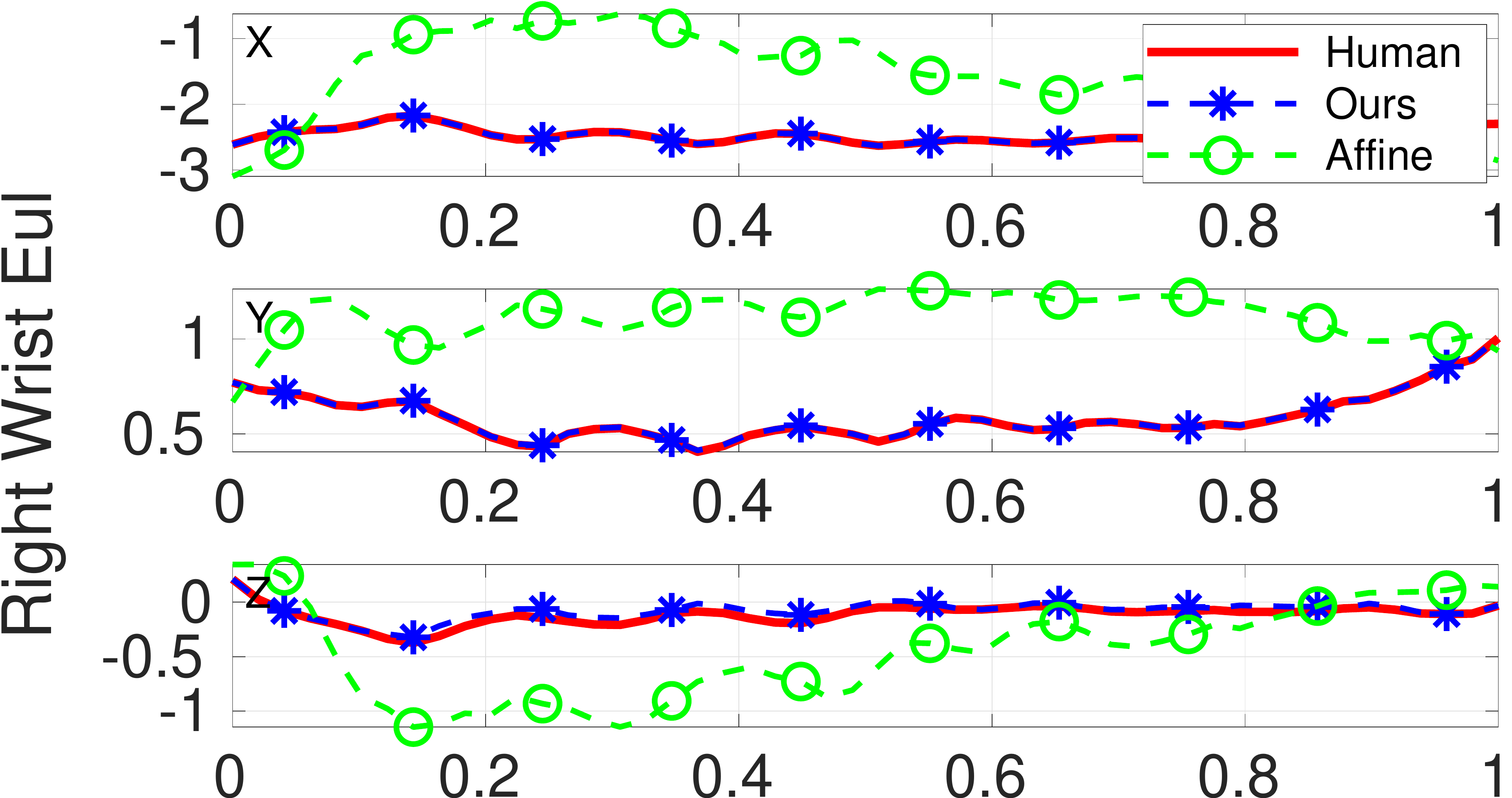}
			\includegraphics[width=0.32\linewidth]{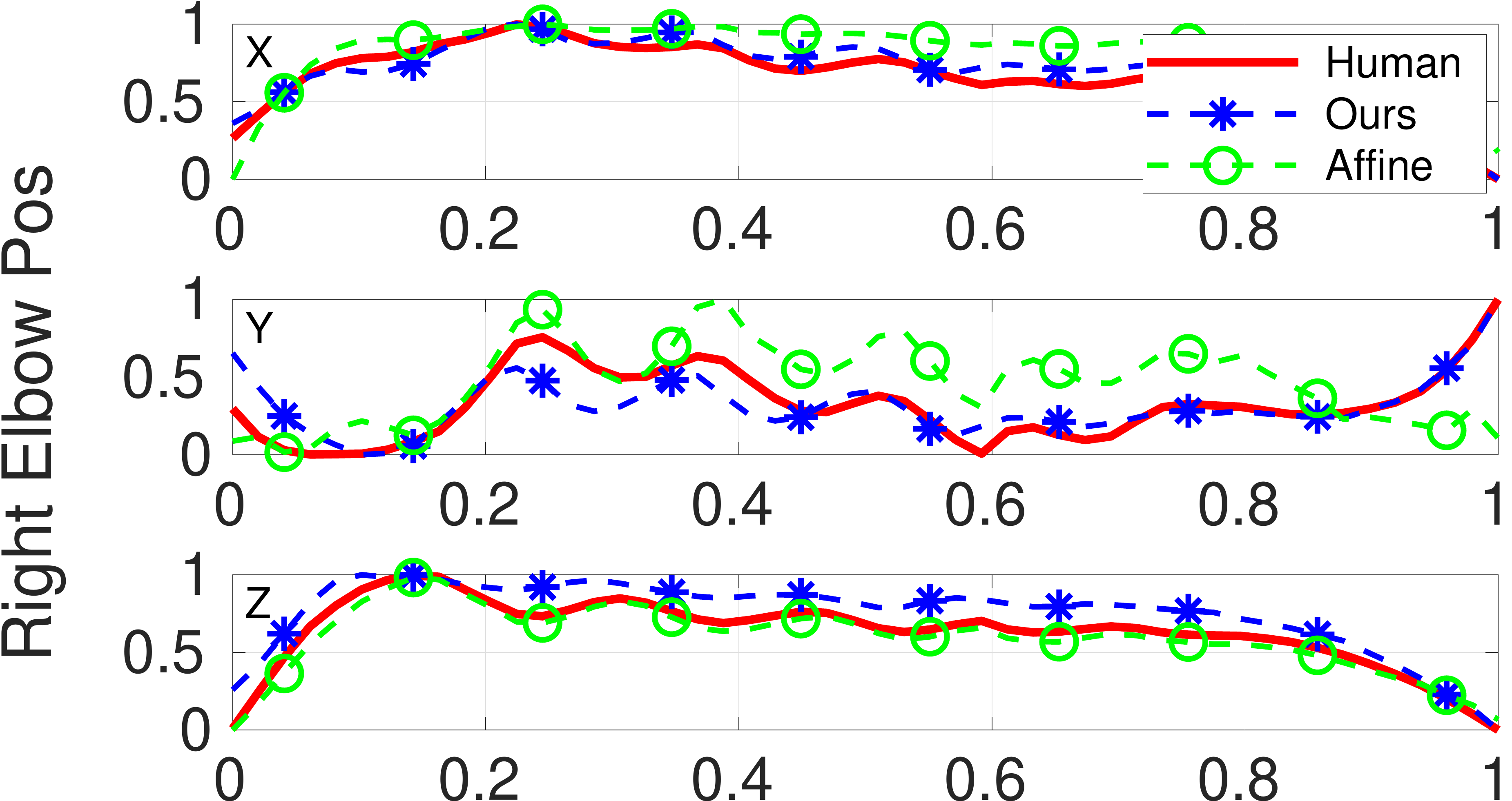}
			\label{fig:r_arm_fengren_1}
		} 
		\subfigure{
			\includegraphics[width=0.32\linewidth]{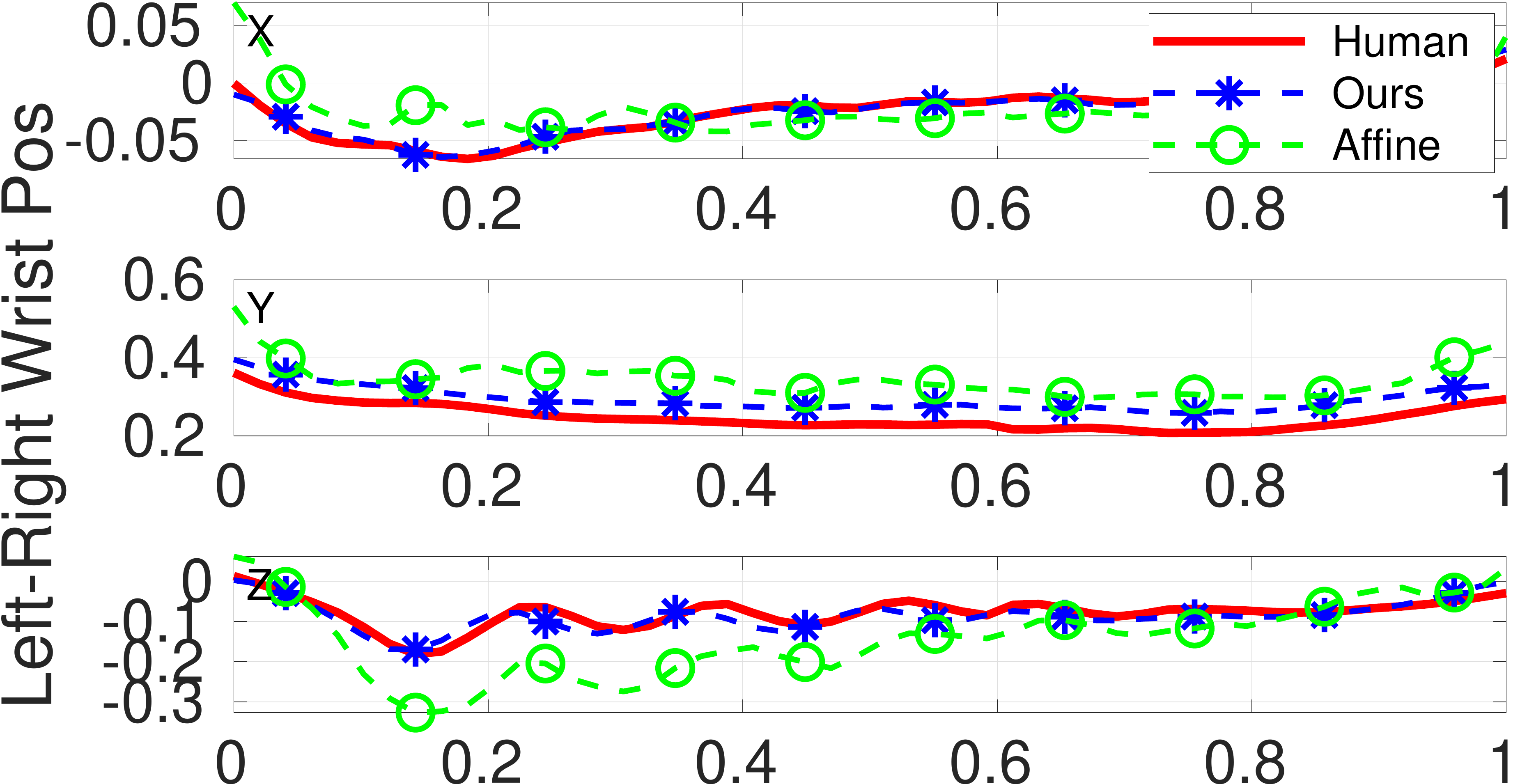}
			\includegraphics[width=0.32\linewidth]{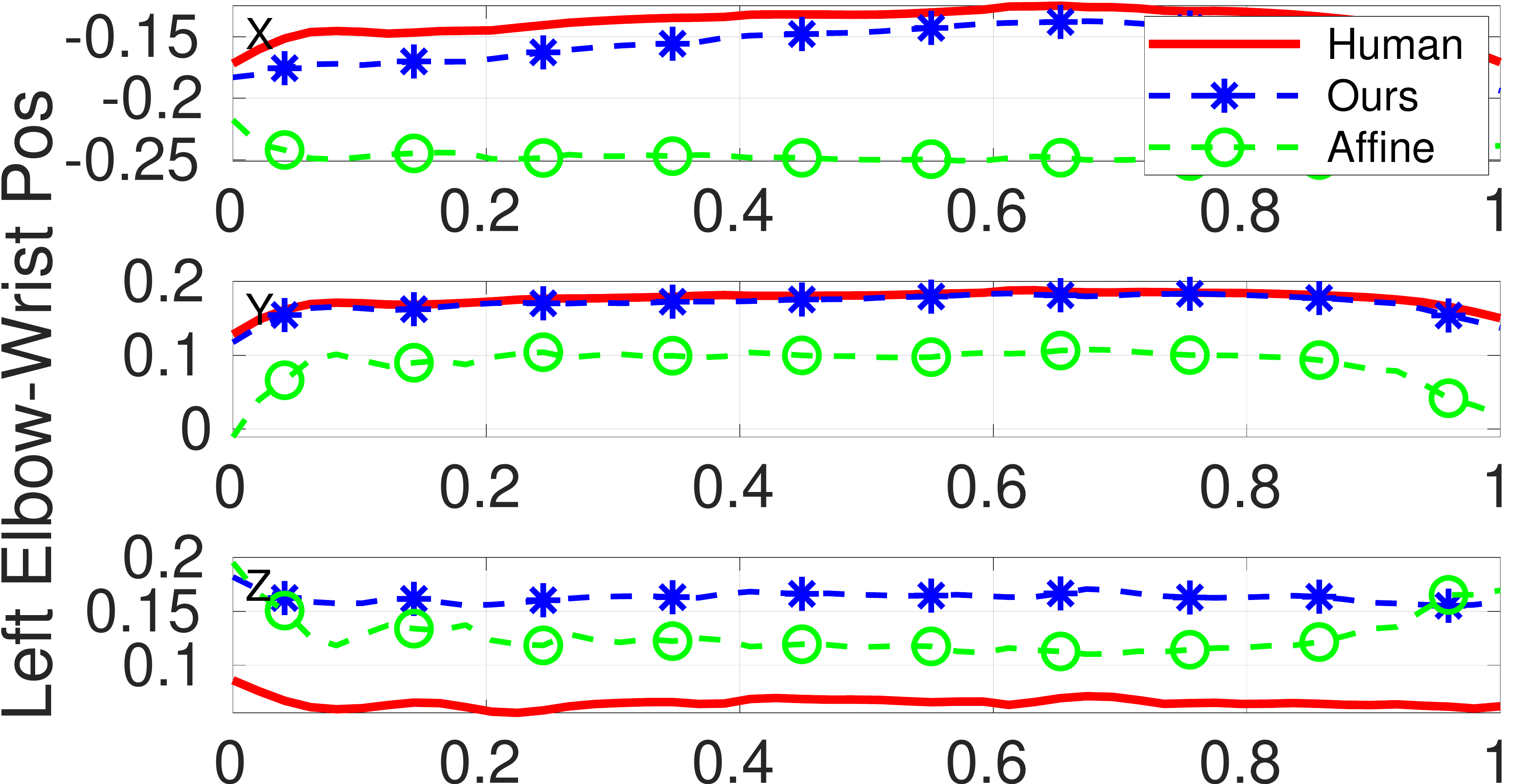}
			\includegraphics[width=0.32\linewidth]{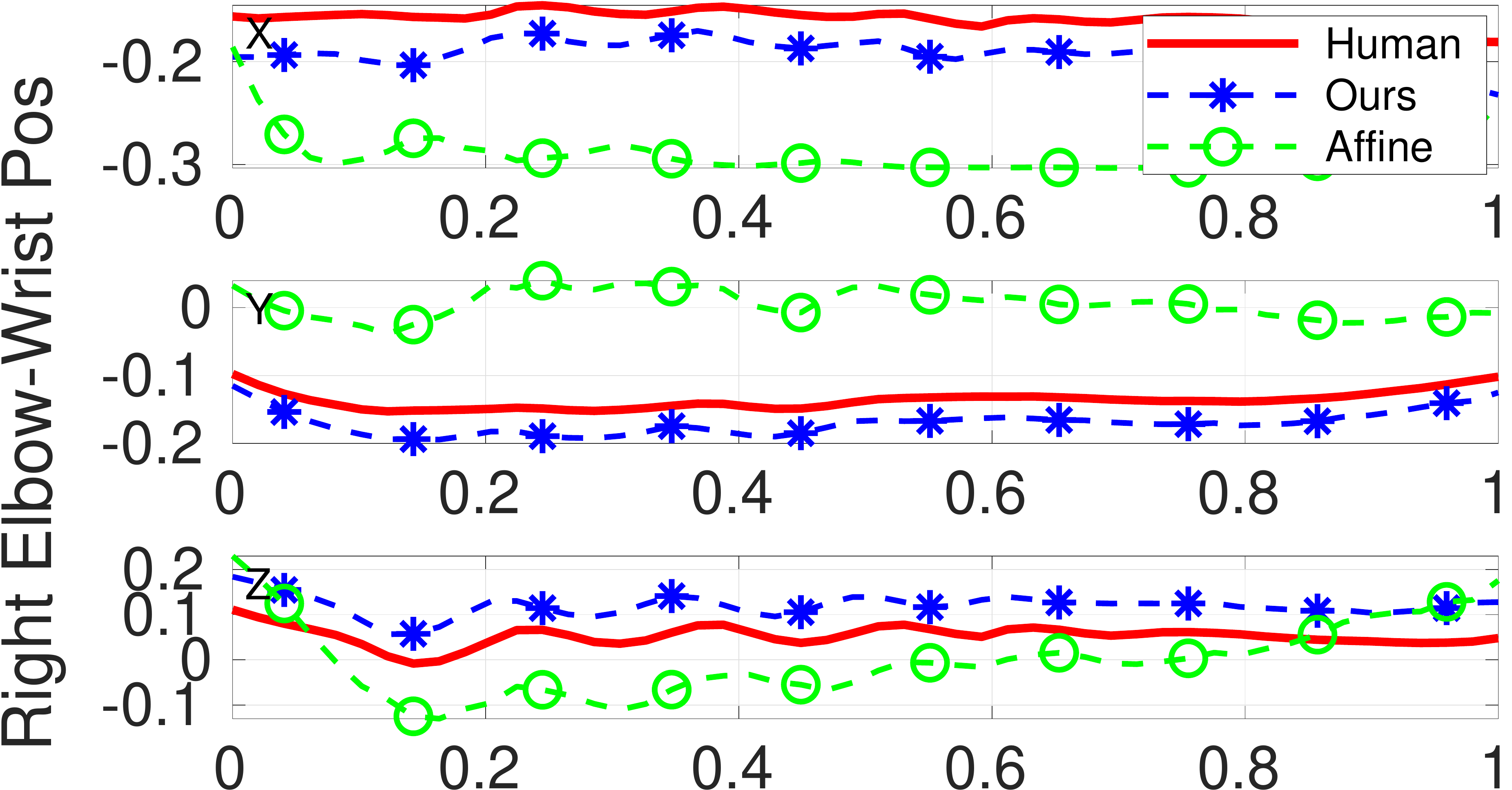}
			\label{fig:rel_fengren_1}
		} 
		\caption{Comparison of feasible retargeted results of $``fengren"$ (sewing) motion. Absolute (normalized to $[0,1]$) and relative position trajectories, as well as orientation trajectories, are displayed. Time is normalized according to the path length.}
		\label{fig:fengren_1_curves}
		\vspace{-5mm}
	\end{figure*}

	\textbf{Qualitative Case Study.} To examine the performance of our method, an example of a motion $``fengren"$ (sewing) is chosen, for which absolute and relative position trajectories, as well as orientation trajectories, are plotted in Fig. \ref{fig:fengren_1_curves}.
	In the first two rows, all of the absolute trajectories are close to those of the original motion, indicating that our method preserves the original pattern of human demonstration. For the relative trajectories, there are slight differences in some components, e.g. z-axis of $LEW$ trajectory, but the overall path shapes are well preserved which can be verified by Table \ref{table:frechet_dists3}. We argue that such translational deviation in relative trajectories are allowed and even necessary because of differences in body structure and size between humans and robots. For example, a pause gesture may require a larger distance between a robot's wrists if its hands are bigger in size than those of the human demonstrator, so as to be a feasible motion for robot execution.

	\newcommand{\spsize}{0.18}
	\begin{figure}[htb]
		\centering
		\subfigure[Human demonstration]{
			\includegraphics[width=\spsize\linewidth]{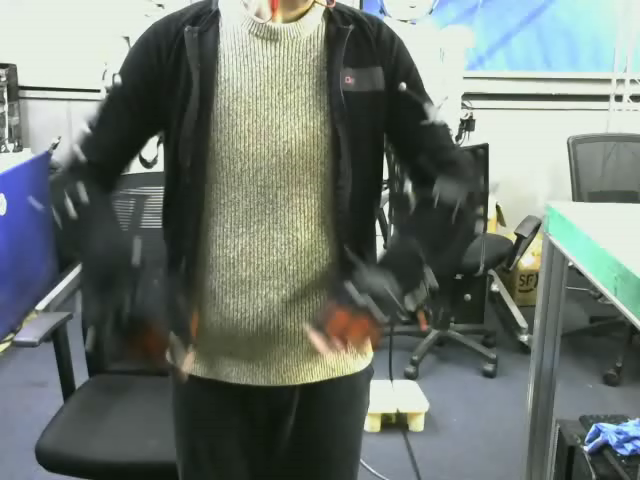}
			\includegraphics[width=\spsize\linewidth]{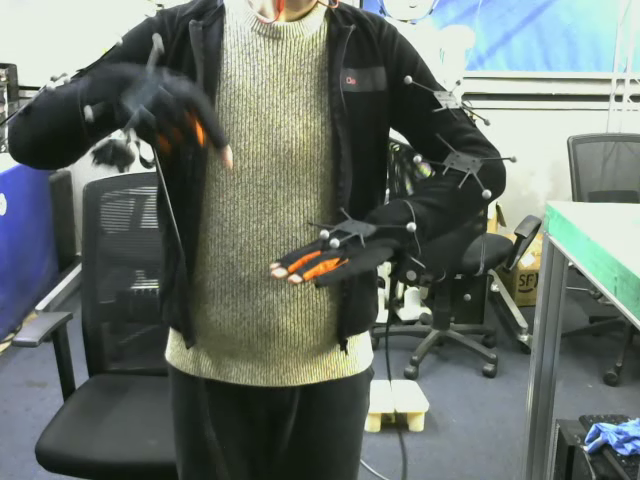}
			\includegraphics[width=\spsize\linewidth]{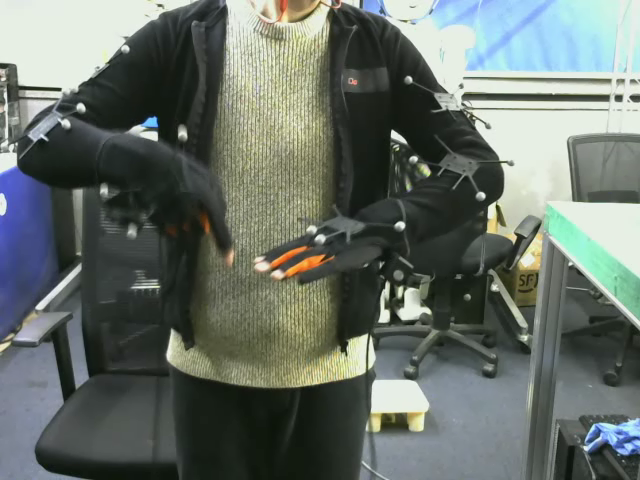}
			\includegraphics[width=\spsize\linewidth]{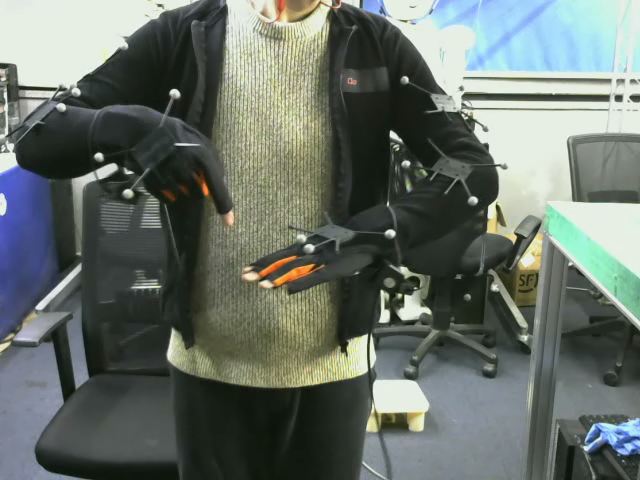}
			\includegraphics[width=\spsize\linewidth]{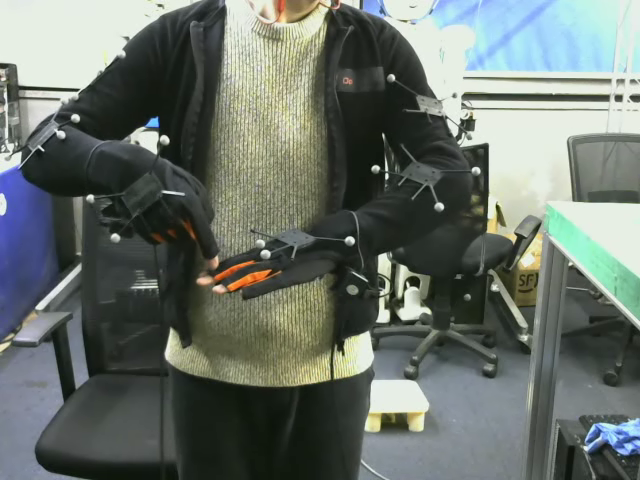}
			\label{fig:fengren_1_snapshots_original}
		}
		\subfigure[Pure IK results.]{
			\includegraphics[width=\spsize\linewidth]{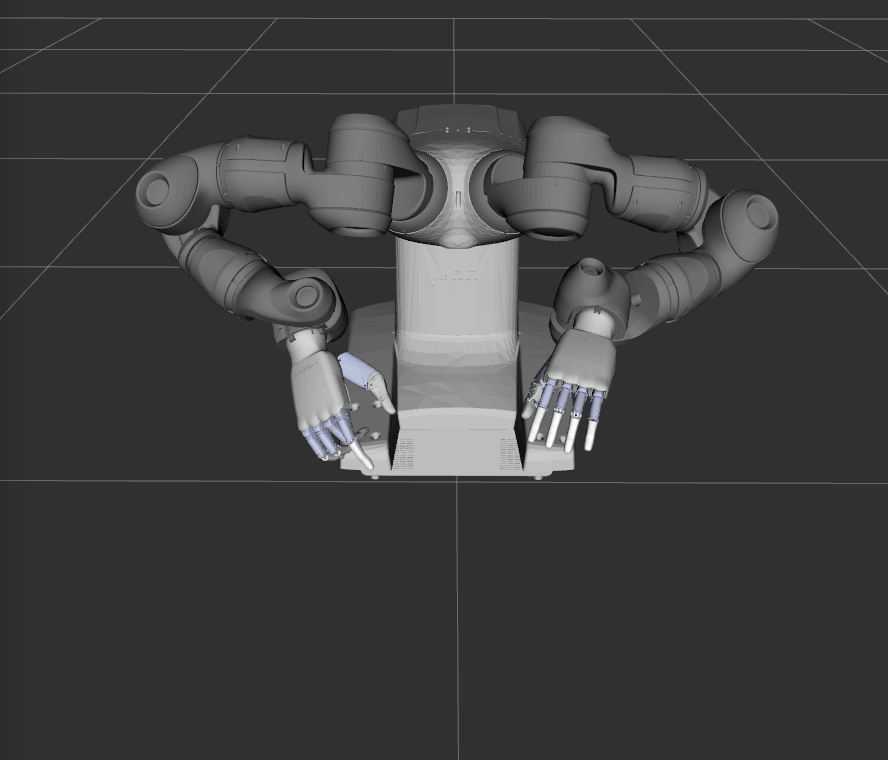}
			\includegraphics[width=\spsize\linewidth]{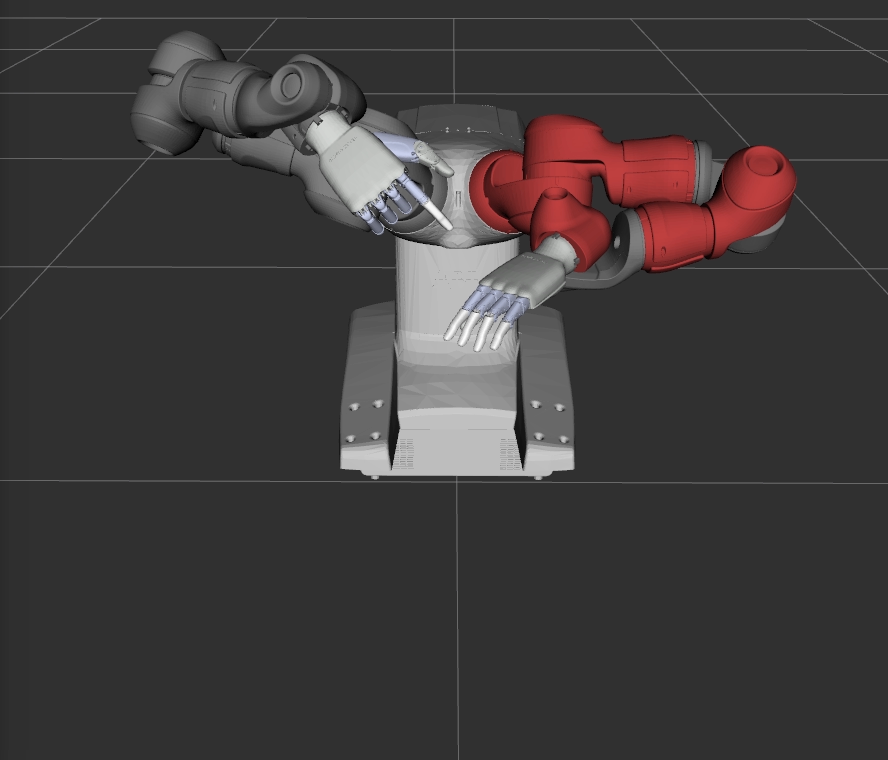}
			\includegraphics[width=\spsize\linewidth]{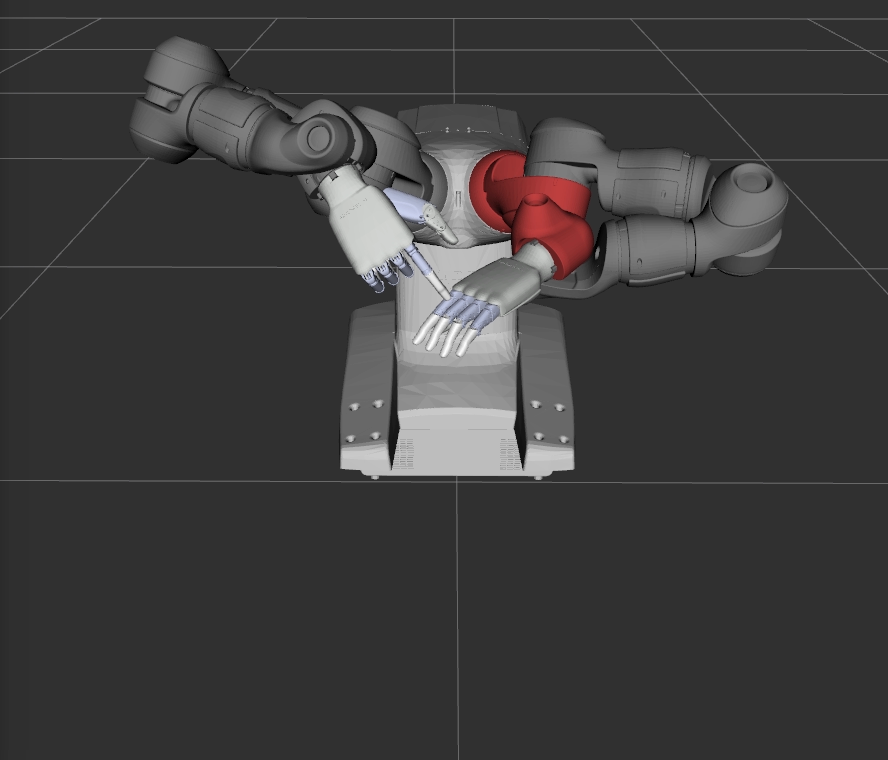}
			\includegraphics[width=\spsize\linewidth]{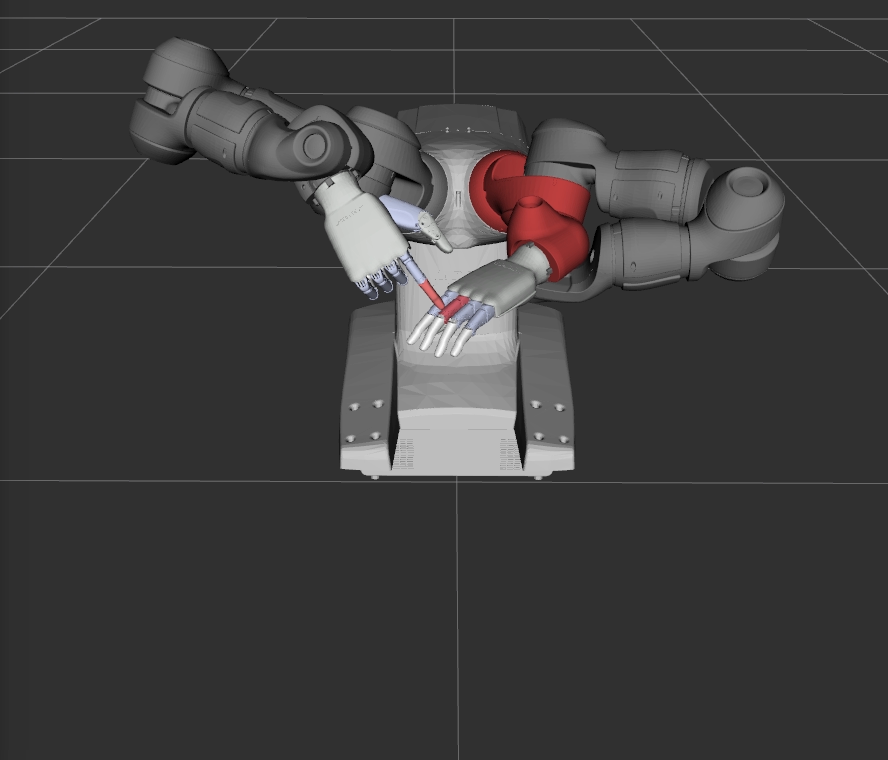}
			\includegraphics[width=\spsize\linewidth]{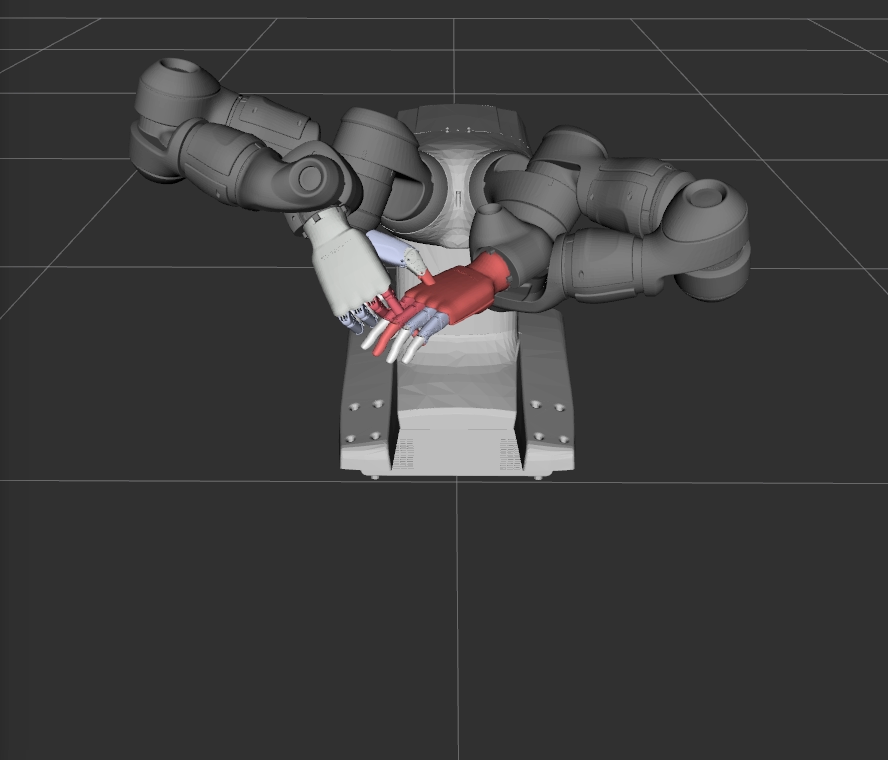}
			\label{fig:fengren_1_snapshots_pure_ik}
		}
		\subfigure[Position Scaling results.]{
			\includegraphics[width=\spsize\linewidth]{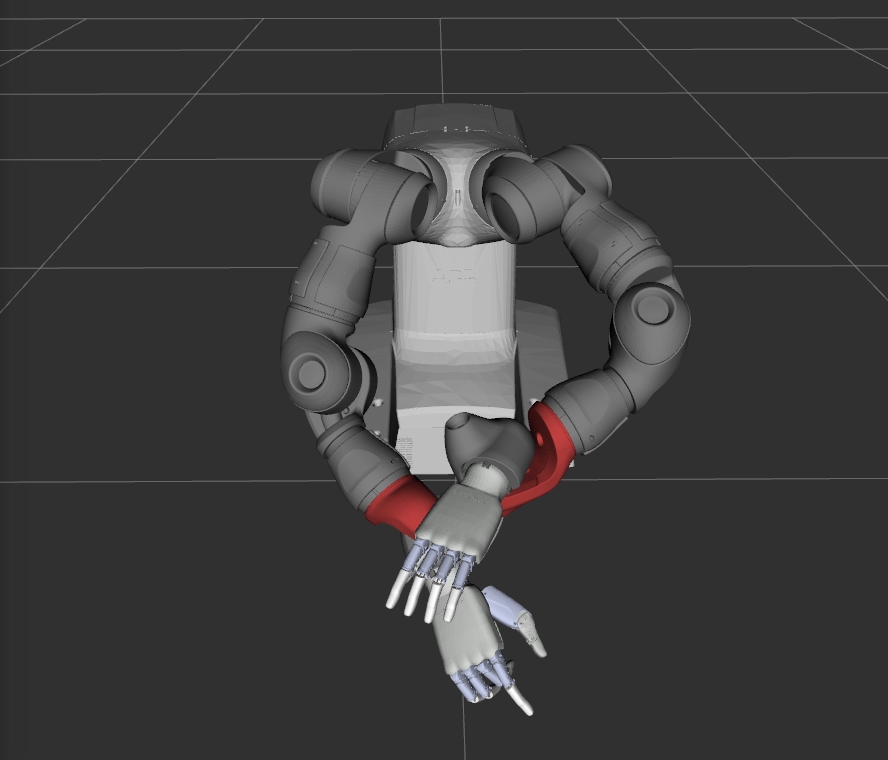}
			\includegraphics[width=\spsize\linewidth]{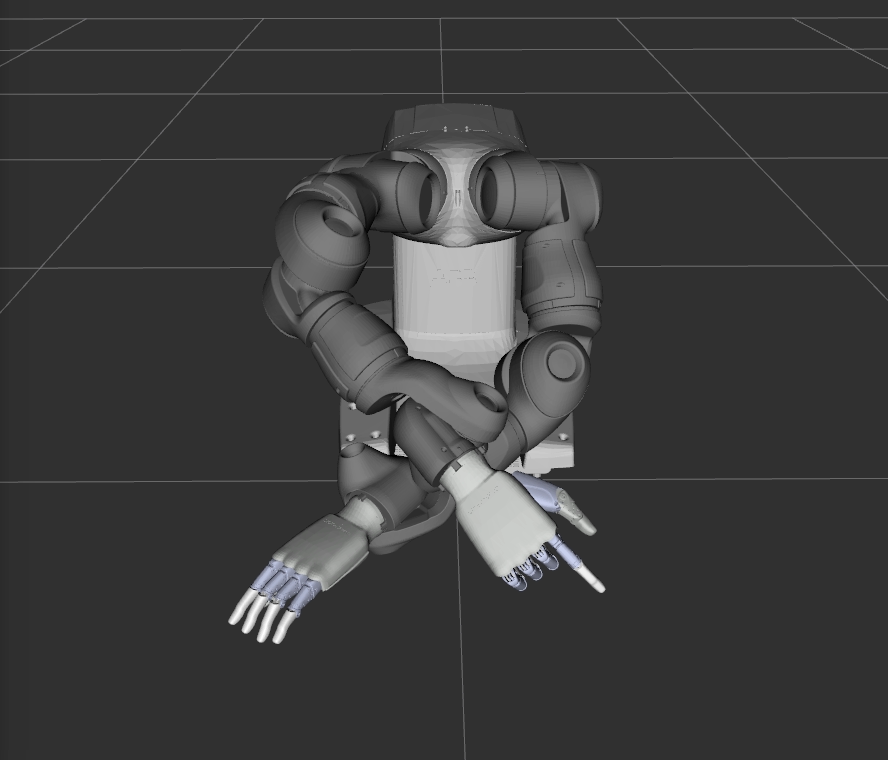}
			\includegraphics[width=\spsize\linewidth]{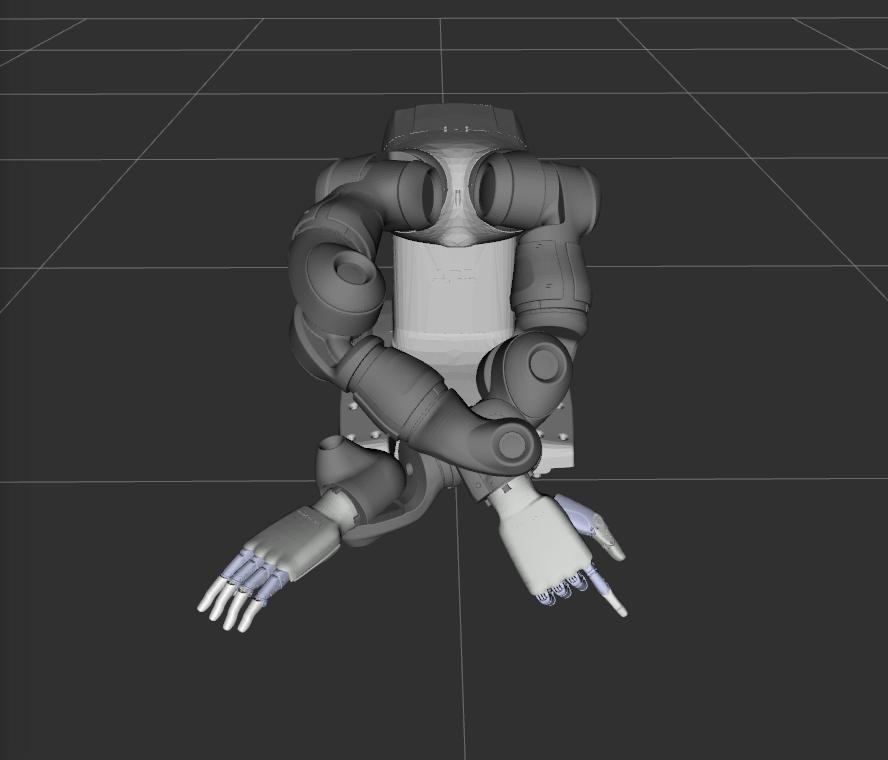}
			\includegraphics[width=\spsize\linewidth]{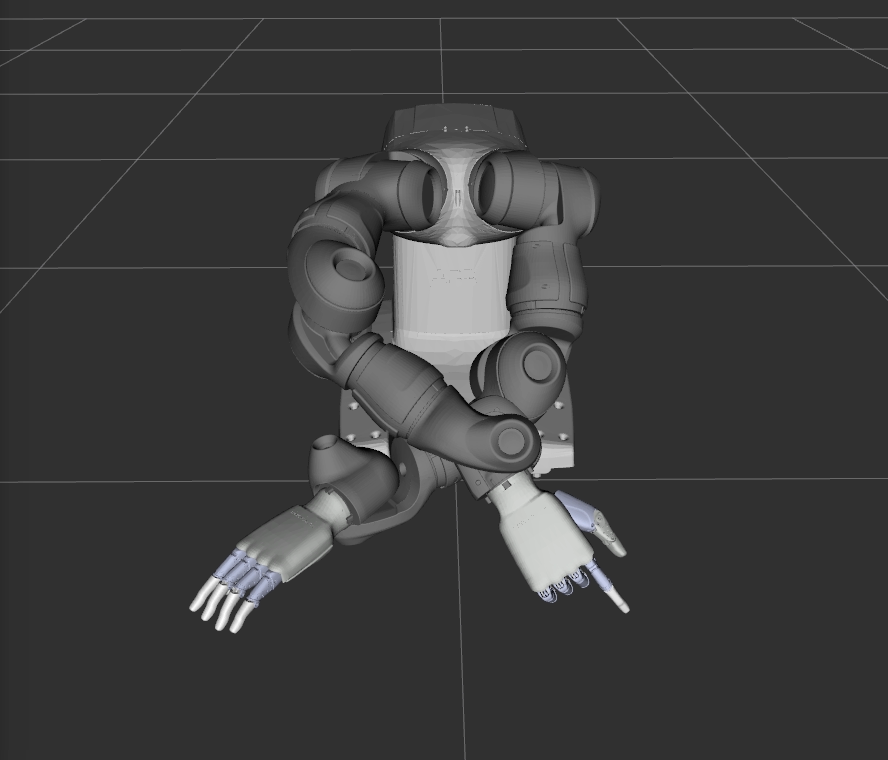}
			\includegraphics[width=\spsize\linewidth]{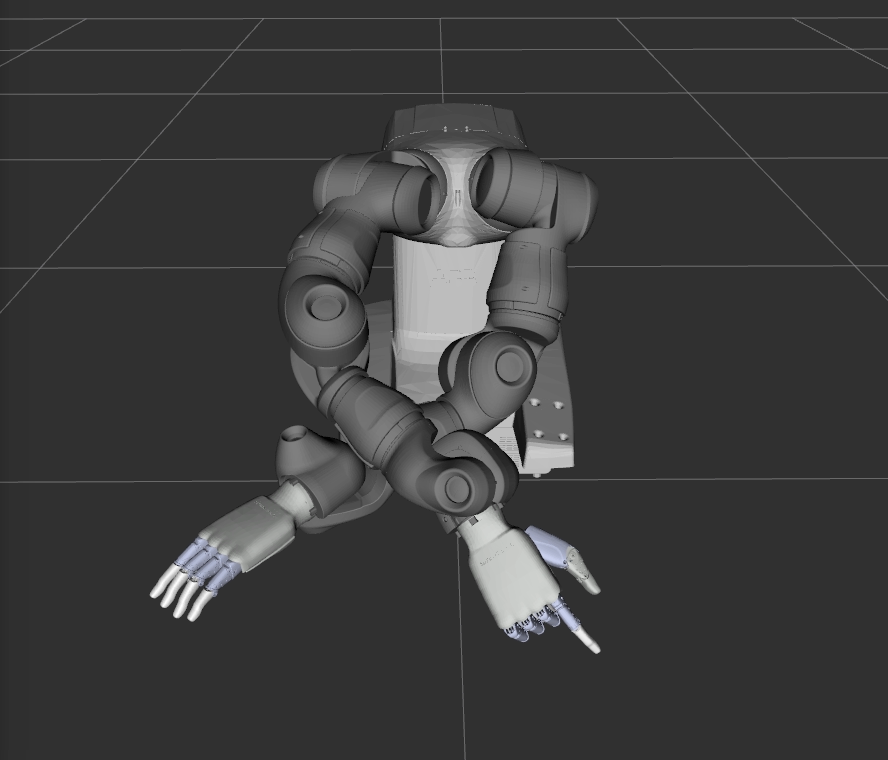}
			\label{fig:fengren_1_snapshots_pos_scaling}
		}
		\subfigure[Affine Deformation results.]{
			\includegraphics[width=\spsize\linewidth]{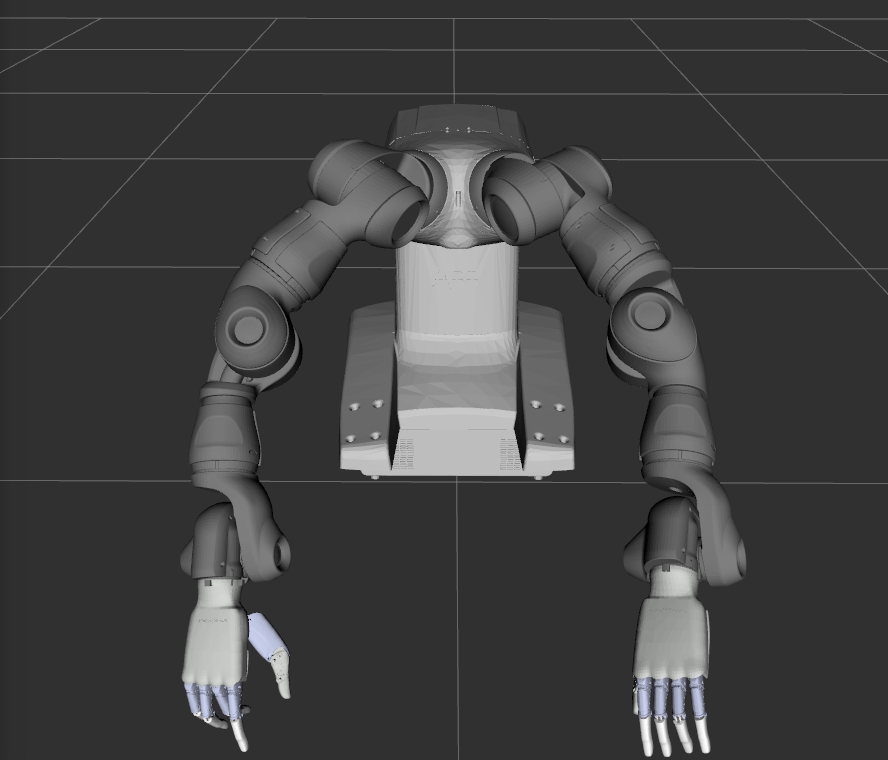}
			\includegraphics[width=\spsize\linewidth]{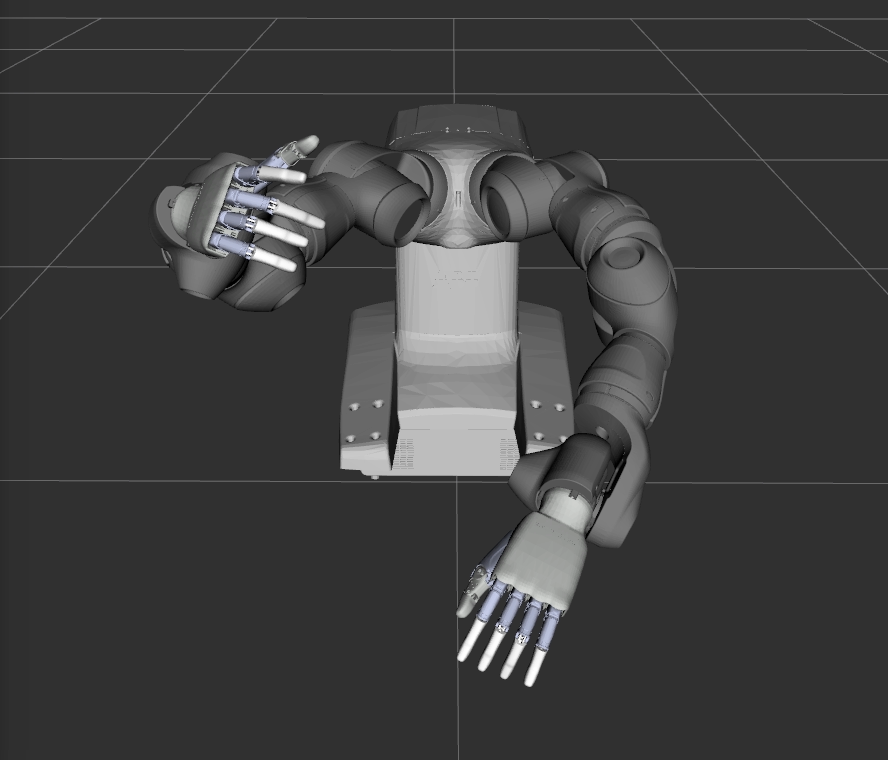}
			\includegraphics[width=\spsize\linewidth]{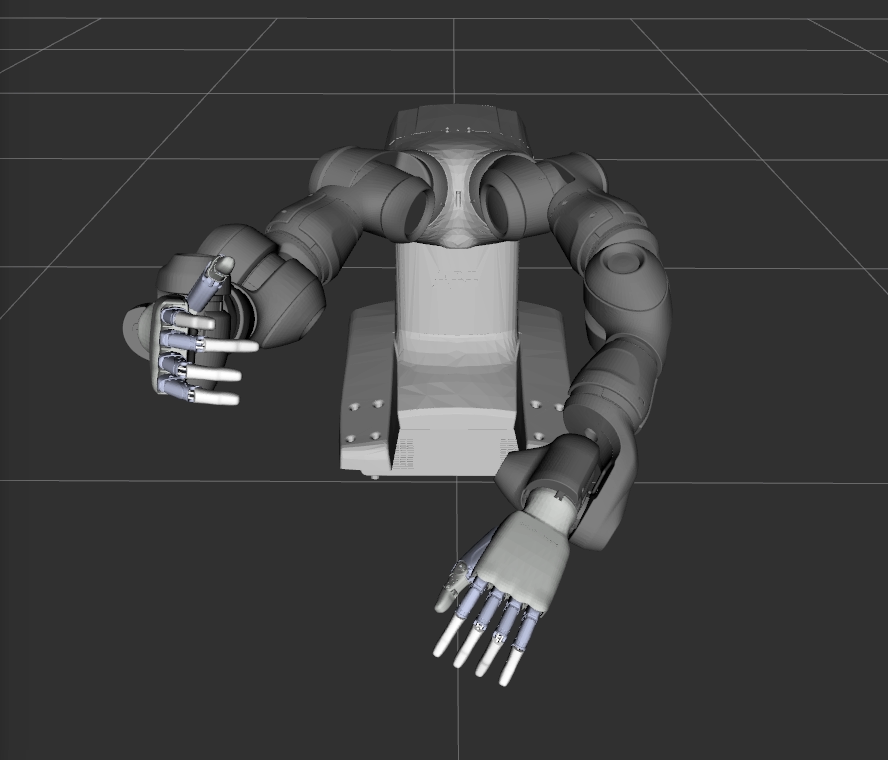}
			\includegraphics[width=\spsize\linewidth]{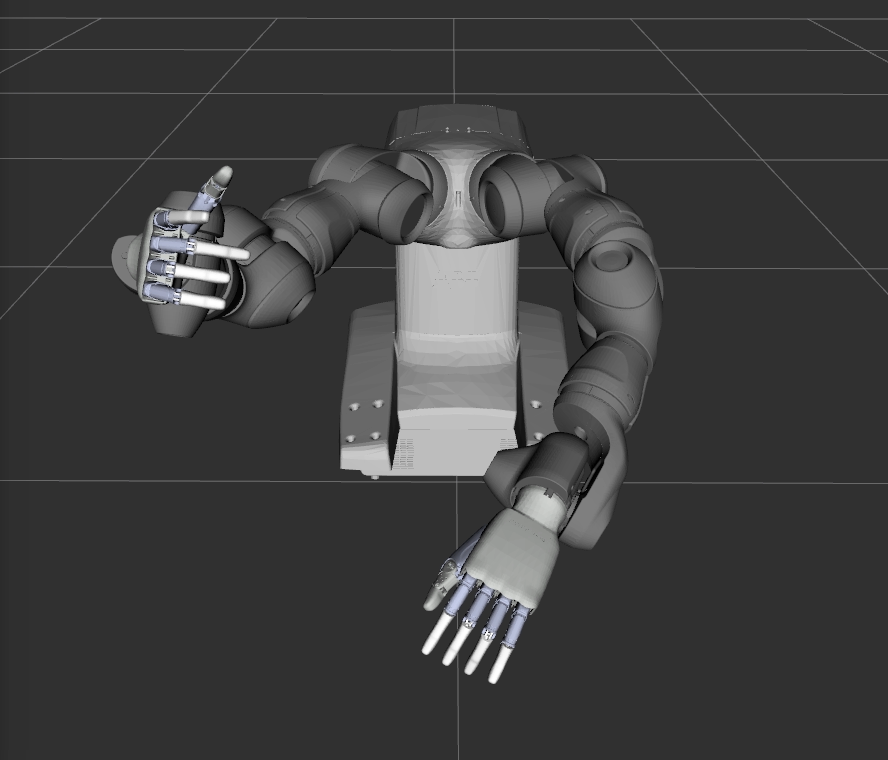}
			\includegraphics[width=\spsize\linewidth]{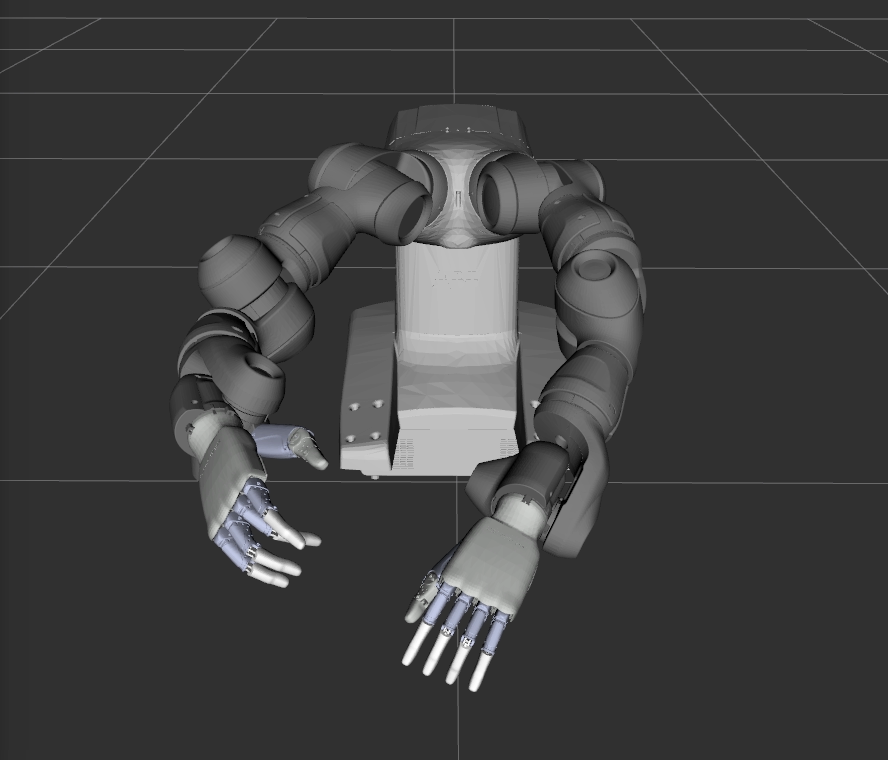}
			\label{fig:fengren_1_snapshots_hujin}
		}
		\subfigure[Our results.]{
			\includegraphics[width=\spsize\linewidth]{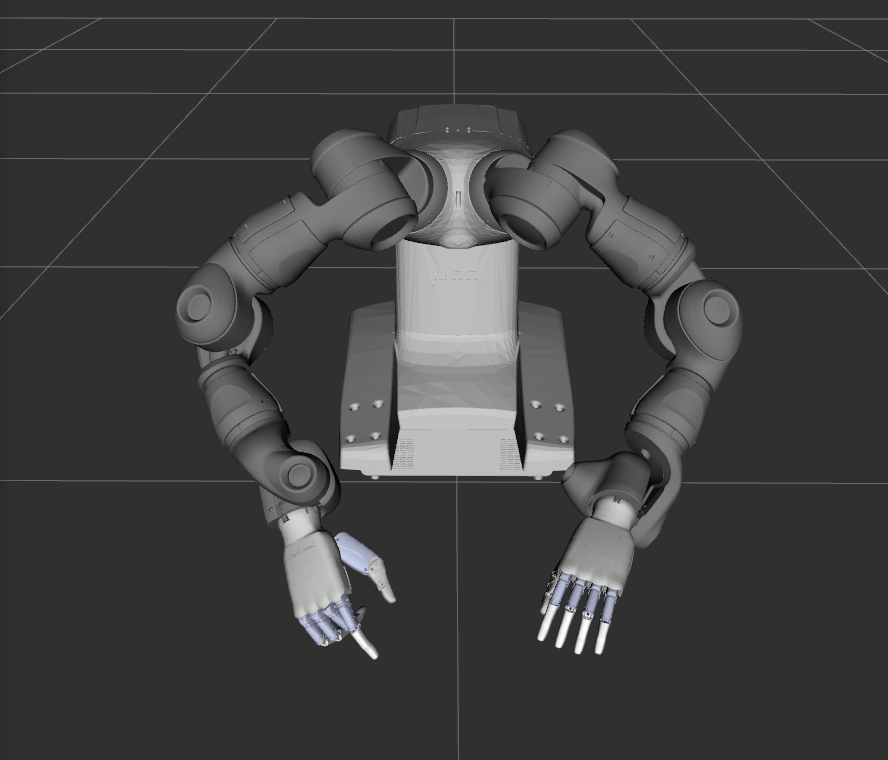}
			\includegraphics[width=\spsize\linewidth]{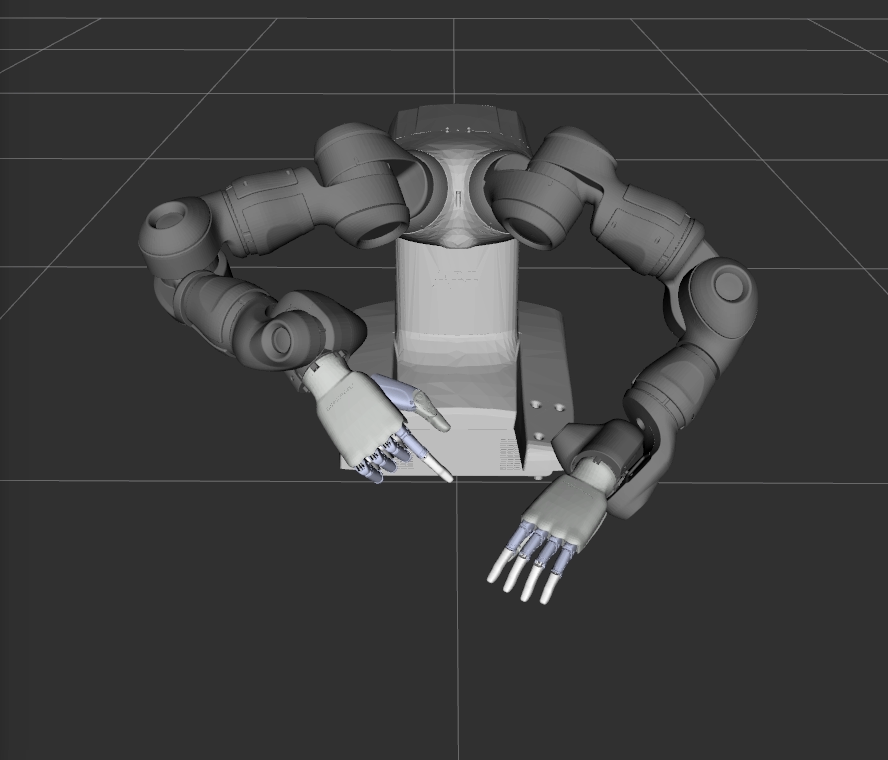}
			\includegraphics[width=\spsize\linewidth]{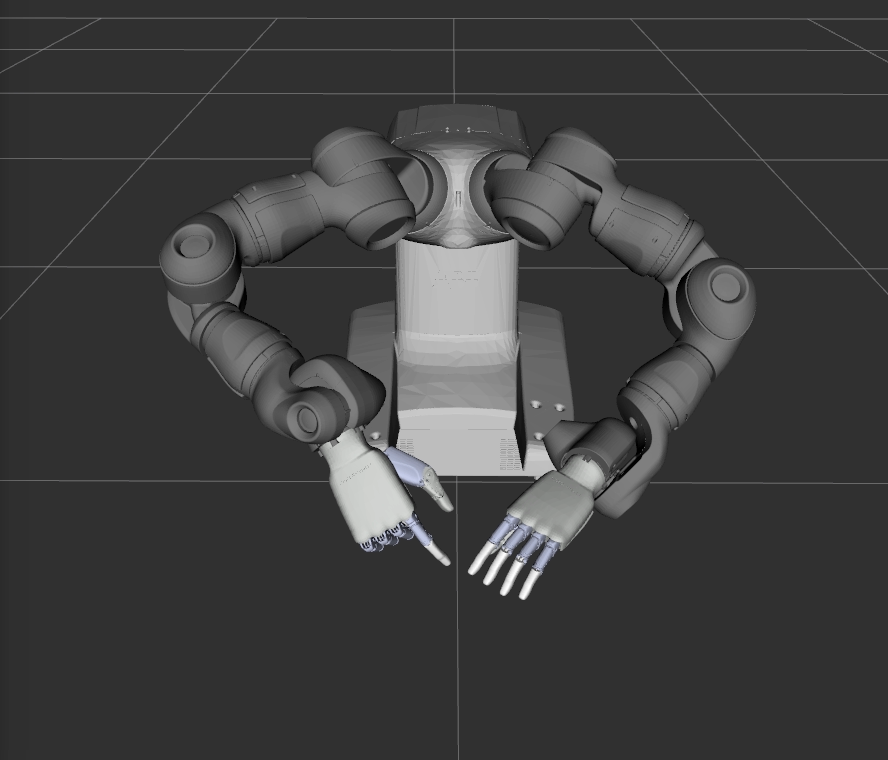}
			\includegraphics[width=\spsize\linewidth]{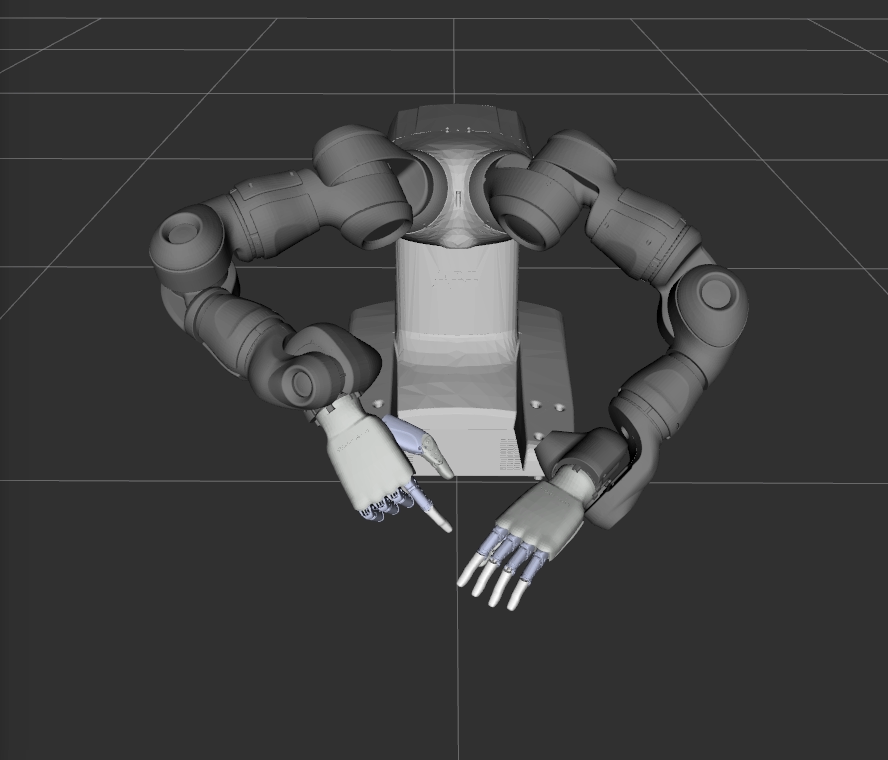}
			\includegraphics[width=\spsize\linewidth]{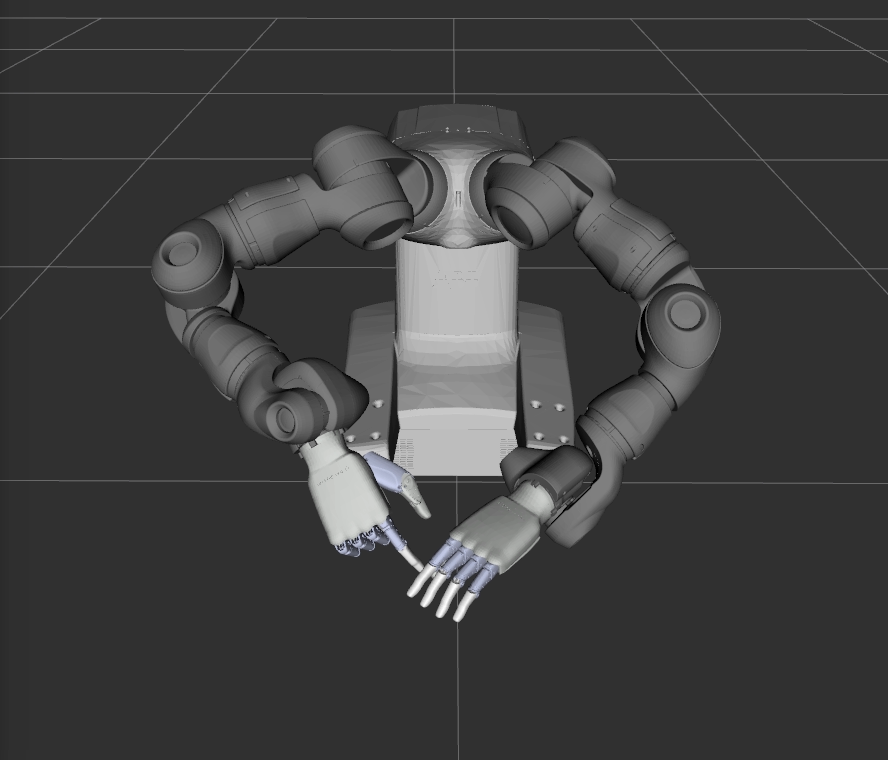}
			\label{fig:fengren_1_snapshots_ours}
		}
		\caption{Snapshots of $``fengren"$ (sewing) motion.}
		\label{fig:fengren_1_snapshots}
		\vspace{-8mm}
	\end{figure}

	Snapshots of the original and retargeted motions are displayed in Fig. \ref{fig:fengren_1_snapshots}, which show that our retargeted motion better resembles the original one. The PureIK method yields colliding results due to different body size. The position scaling approach even results in serious misalignment between both wrists. And the affine transformation method does not preserve the wrist's orientation well.

	\section{Conclusions}
	
	In this paper, we propose a novel DMP-based motion retargeting method for transferring human's complex dual-arm motions such as sign language motions to robots. We test our method with a dual-arm robot with multi-fingered hands, an arm-hand system with 26 DOFs in total, for retargeting a subset of words chosen from Chinese Sign Language. We also conduct comparison experiments to validate its ability to effectively preserve the motion rhythm and coordination pattern of human demonstrations.

	

\bibliographystyle{IEEEtran} 

\end{document}